\documentclass{article}

\usepackage[a4paper, total={6in, 9in}]{geometry}
\usepackage[round]{natbib}

\usepackage[utf8]{inputenc} 
\usepackage[T1]{fontenc}    
\usepackage{hyperref}       
\usepackage{url}            
\usepackage{booktabs}       
\usepackage{amsfonts}       
\usepackage{nicefrac}       
\usepackage{microtype}      
\usepackage[dvipsnames]{xcolor}         

\usepackage{microtype}
\usepackage{graphicx}
\usepackage{subfigure}
\usepackage{booktabs} 
\usepackage{stmaryrd}
\usepackage{amsmath,amssymb,amsthm}
\usepackage{dsfont}
\usepackage{tikz}
\usepackage{pifont}
\usepackage{enumitem}
\usepackage{sidecap}
\usepackage{algorithm}
\usepackage{algorithmic}
\usepackage{amsmath}
\usepackage{wrapfig}
\usepackage{changes}
\usepackage{xspace}
\usepackage{authblk}
\usepackage{minitoc}
\definecolor{darkcyan}{rgb}{0.0, 0.55, 0.55}
\definecolor{azure}{rgb}{0.0, 0.5, 1.0}
\definecolor{cadmiumred}{rgb}{0.89, 0.0, 0.13}
\hypersetup{ %
colorlinks=true,
linkcolor=cadmiumred,
citecolor=azure,
filecolor=azure,
urlcolor=azure}

\newcommand{\enc}[1]{{\color{darkcyan}#1}}

\DeclareRobustCommand{\eg}{e.g.,\@\xspace}
\DeclareRobustCommand{\ie}{i.e.,\@\xspace}

\newcommand{\linucb}{\textsc{LinUCB}\xspace}
\newcommand{\oful}{\textsc{OFUL}\xspace}
\newcommand{\algo}{\textsc{HELBA}\xspace}
\newcommand{\secoful}{\algo}
\newcommand{\secofulls}{\algo}
\newcommand{\rsoful}{\textsc{RSOFUL}\xspace}
\newcommand{\rsofultr}{\textsc{RSOFUL-Tr}\xspace}

\newcommand{\wt}[1]{\widetilde{#1}}
\newcommand{\wh}[1]{\widehat{#1}}
\newcommand{\wb}[1]{\overline{#1}}

\newtheorem{theorem}{Theorem}
\newtheorem{proposition}[theorem]{Proposition}

\newtheorem{assumption}[theorem]{Assumption}
\newtheorem{corollary}[theorem]{Corollary}
\newtheorem{lemma}[theorem]{Lemma}

\newtheorem*{theorem*}{Theorem}
\newtheorem*{proposition*}{Proposition}
\newtheorem*{corollary*}{Corollary}

\def\showcomments{0}
\ifnum\showcomments=1
\definecolor{cadmiumred}{rgb}{0.89, 0.0, 0.13}
\newcommand{\todoe}[1]{\todo[color=orange!30!yellow!10, inline]{\small E: #1}}
\newcommand{\todoeout}[1]{\todo[color=orange!30!yellow!10]{\scriptsize E: #1}}
\newcommand{\todomp}[1]{\todo[color=Green!10, inline]{\small MP: #1}}
\newcommand{\todompout}[1]{\todo[color=Green!10]{\scriptsize MP: #1}}
\newcommand{\todovp}[1]{\todo[color=Magenta!10, inline]{\small VP: #1}}
\newcommand{\todovpout}[1]{\todo[color=Magenta!10]{\scriptsize VP: #1}}
\newcommand{\todohc}[1]{\todo[color=blue!10, inline]{\small HC: #1}}
\newcommand{\todohcout}[1]{\todo[color=blue!10]{\scriptsize HC: #1}}
\else
\definecolor{violet}{rgb}{0, 0, 0}
\newcommand{\todoe}[1]{}
\newcommand{\todoeout}[1]{}
\newcommand{\todomp}[1]{}
\newcommand{\todompout}[1]{}
\newcommand{\todovp}[1]{}
\newcommand{\todovpout}[1]{}
\newcommand{\todohc}[1]{}
\newcommand{\todohcout}[1]{}
\fi

\usepackage[toc,page,header]{appendix}
\usepackage{minitoc}

\title{Encrypted Linear Contextual Bandit}

\author[1,2]{Evrard Garcelon}
\affil[1]{Meta AI}
\affil[2]{CREST, ENSAE}
\author[2]{Vianney Perchet}
\author[1]{Matteo Pirotta}
\date{}

\begin{document}

%

%

\doparttoc 
\faketableofcontents 

\maketitle

\begin{abstract}
  Contextual bandit is a general framework for online learning in sequential decision-making problems that has found application in a wide range of domains, including recommendation systems, online advertising, and clinical trials.
  A critical aspect of bandit methods is that they require to observe the contexts --i.e., individual or group-level data-- and rewards in order to solve the sequential problem.
  The large deployment in industrial applications has increased interest in methods that preserve the users' privacy.
  In this paper, we introduce a privacy-preserving bandit framework based on homomorphic encryption{\color{violet} which allows computations using encrypted data}. The algorithm \textit{only} observes encrypted information (contexts and rewards) and has no ability to decrypt it.
  Leveraging the properties of homomorphic encryption, we show that despite the complexity of the setting, it is possible to {\color{violet} solve linear contextual bandits over encrypted data with a $\widetilde{O}(d\sqrt{T})$ regret bound in any linear contextual bandit problem, while keeping data encrypted.}
\end{abstract}

\section{INTRODUCTION}\label{sec:introduction}
Contextual bandits have become a key part of several applications such as
marketing, healthcare and finance; as they can be used to provide personalized \eg adaptive service~\citep{bastani2020online, sawant2018contextual}.
{\color{violet} In such application, algorithms receives as input users' features, i.e.\ the ``contexts'', to tailor their recommendations. Those features} may disclose sensitive information, as personal  (e.g., age, gender, etc.) or geo-localized features are  commonly used in recommendation systems.
Privacy awareness has increased over years and users are less willing to disclose information and are more and more concerned about how their personal data is used \citep{das2021panel}.
For example, a user may be willing to receive financial investment suggestion but not to share information related to income, deposits, properties owned and other assets. However, without observing this important information about a user, a service provider may not be able to provide meaningful investment guidance to the user. This example extends to many other applications. For instance,
suppose an user is looking for a restaurant nearby, if the provider has no access to even a coarse geo-location, it would not be able to provide meaningful suggestions to the user.
An effective approach 
 to address these concerns is to resort to end-to-end encryption to guarantee that data is readable only by the users~\citep{kattadige2021360norvic}.
In this scenario, the investment company or the service provider observes only an encrypted version of user's information and have no ability to decrypt it.
While this guarantee high level of privacy, it is unclear whether the problem remains learnable and how to design effective online learning algorithms in this secure scenarios.


In this paper, we introduce - and analyze - the setting of encrypted contextual bandit to model the mentioned scenarios.
At each round, a bandit algorithm observes encrypted features (including e.g., geo-location, food preferences, visited restaurants), chooses an action (e.g., a restaurant) and observes an encrypted reward (e.g., user's click), that is used to improve the quality of recommendations.
While it is possible to obtain end-to-end encryption --i.e., the bandit algorithm only observes encrypted information that is not able to decrypt-- using standard encryption methods (e.g., AES, RSA, TripleDES), the provider may no longer be able to provide a meaningful service since may not be able to extract meaningful information from encrypted features. We thus address the following question:

\begin{center}
    \emph{Is it possible to learn with encrypted contexts and rewards? And what is the associated computational and learning cost?}
\end{center}
Homomorphic Encryption~\citep[][HE]{halevi2017homomorphic} is a powerful encryption method that allows to carry out computation of encrypted numbers. While this is a very powerful idea, only a limited number of operations can be performed, notably only addition and/or multiplication. While HE has been largely investigated in supervised learning~\citep{badawi2020alexnet,graham2015fractional}, little is known about online learning. In this paper we aim to look into this direction.
We approach the aforementioned question via HE and from a theoretical point-of-view. We consider the case of linear rewards and investigate the design of a ``secure'' algorithm able to achieve sub-linear regret in this setting. There are several challenges in the design of bandit algorithms that makes the application of HE techniques not easy. First, it is not obvious that all the operations required by a bandit algorithm (notably optimism) can be carried out only through additions and multiplications. Second, errors or approximations introduced by the HE framework to handle encrypted data may compound and prevent to achieve provably good performance. Finally, a careful algorithmic design is necessary to limit the total number of HE operations, which are computationally demanding.

\textbf{Contributions.}
Our main contributions can be summarized as follows:
    \textbf{1)} We introduce and formalize the problem of secure contextual bandit with homomorphic encryption.
    \textbf{2)} We provide the first bandit algorithm able to learn over encrypted data in contextual linear bandits, a standard framework that allows us to describe and address all the challenges in leveraging HE in online learning. Leveraging optimism~\citep[\eg][]{abbasi2011improved} and HE, we introduce \algo which balances security, approximation error due to HE and computational cost to achieve a $\wt{O}(\sqrt{T})$ regret bound. This shows that i) it is possible to learn \emph{online} with encrypted information; ii) preserving users’ data security has negligible impact on the learning process. This is a large improvement w.r.t.\ $\varepsilon$-LDP which has milder security guarantees and where the best known bound is $\wt{O}(T^{3/4}/\varepsilon)$.
    \textbf{3)}  We discuss practical limitations of HE and ways of improving the efficiency of the proposed algorithm, {\color{violet} mainly how the implementation of some procedures can speed up computations and allow to scale dimension of contexts}. We report preliminary numerical simulations confirming the theoretical results.\todovpout{VP: we should put more emphasis on the tractability and all the complexity questions}

\textbf{Related work.}
To prevent information leakage, the bandit literature has mainly focused
on Differential Privacy (DP)~\citep[e.g.,][]{shariff2018differentially, TossouD16}.
While standard $(\epsilon,\delta)$-DP enforces statistical diversity of the output of an algorithm, it does not provide guarantees on the security of user data that can be accessed directly by the algorithm.
A stronger privacy notion, called local DP, requires data being privatized before being accessed by the algorithm.
While it may be conceptually similar to encryption, i) it does not provide the same security guarantee as encryption (having access to a large set of samples may allow some partial denoising~\cite{cheu2019manipulation}); and ii) it has a large impact on the regret of the algorithm. For example, \cite{zheng2021locally} recently analyzed $\varepsilon$-LDP in contextual linear bandit and derived an algorithm with $\wt{O}(T^{3/4}/\varepsilon)$ regret bound to be compared with a $\wt{O}(\sqrt{T})$ regret of non-private algorithms. Homomorphic Encryption~\citep[e.g.][]{halevi2017homomorphic} has  only been merely used  to encrypt rewards in bandit problems \citep{ciucanu2020secure, ciucanu2019secure}, but in some  inherently simpler setting than the setting considered here (see App.~\ref{app:relwork}).

\section{HOMOMORPHIC ENCRYPTION}\label{sec:he}
Homomorphic Encryption~\citep{halevi2017homomorphic} is a \emph{probabilistic encryption method} that enables an untrusted party to perform some computations (addition and/or multiplication)
on encrypted data. Formally, given  two original messages $m_{1}$ and $m_{2} \in \mathbb{R}$,
the addition (resp.\ multiplication) of their encrypted versions (called ciphertexts) is equal to the encryption of their sum $m_{1} + m_{2}$ (resp.\  $m_1 \times m_2$), hence the name ``homomorphic''.\footnote{ Most schemes also support Single Instruction Multiple Data (SIMD), i.e., the same operation on multiple data points in parallel.}
We consider a generic homomorphic schemes that generate a public key
$\textbf{pk}$ (distributed widely and used to encrypt messages), and  private keys
$\textbf{sk}$ (used for decryption of encrypted messages).
This private key is, contrary to the public key, obviously assumed to be kept private.

More precisely, we shall consider \textit{Leveled Fully Homomorphic} encryption (LFHE)
schemes for real numbers. This type of schemes supports both {\color{violet} additions and multiplications} 
but only for a fixed and finite number of operations,
referred to as the \textsl{depth}.
This limitation is a consequence of HE's probabilistic approach.
{\color{violet} Although noisy encryption allows to achieve high security, after a certain number of operations the data is drown in the noise ~\citep[e.g.,][]{albrecht2015concrete}, resulting in an indecipherable
ciphertext (the encrypted message). In most LFHE schemes, the depth is the maximum number of operations possible before losing the ability to decrypt the message. Often
multiplications have a significantly higher noise growth than addition and the depth refers to the maximum number of multiplication between ciphertexts possible. }
The security of a LFHE schemes is defined by $\kappa\in\mathbb{N}$, usually $\kappa \in \{128, 192, 256\}$. A $\kappa$-bit level of security means that an attacker has to perform roughly $2^{\kappa}$ operations to break the encryption scheme, i.e., to decrypt a ciphertext without the secret key.

Formally, an LFHE scheme is defined by:
\begin{itemize}[leftmargin=.1in,topsep=-2.5pt,itemsep=1pt,partopsep=0pt, parsep=0pt]
    \item \textsl{A key generator function}  \textit{KeyGen}$(N, D, \kappa)$:  takes as input the maximum depth $D$ (e.g., max.\ number of multiplications), a security parameter $\kappa$ and the degree $N$ of polynomials used as ciphertexts (App.~\ref{app:ckks}).
    It outputs a secret key $\textbf{sk}$ and a public key $\textbf{pk}$ .
    \item \textsl{An encoding function} $\textit{Enc}_\textbf{pk}(m)$: encrypts the message $m \in \mathbb{R}^d$ with the public key $\textbf{pk}$. The output is a ciphertext $\textbf{ct}$, a representation of $m$ in the space of complex polynomials of degree $N$.
    \item  \textsl{A decoding function} $\textit{Dec}_\textbf{sk}(\textbf{ct})$: decrypts the ciphertext $\textbf{ct}$ of $m\in \mathbb{R}^{d}$ using the secret key $\textbf{sk}$ and outputs message $m$.
    \item \textsl{An additive operator}  \textit{Add}$(\textbf{ct}_{1},\textbf{ct}_{2})$: for ciphertexts $\textbf{ct}_{1}$ and $\textbf{ct}_{2}$ of messages $m_{1}$ and $m_{2}$, it outputs ciphertext $\textbf{ct}_{add}$ of $m_{1} + m_{2}$:
    $\textit{Dec}_\textbf{sk}\Big(\textit{Add}\big(\textit{Enc}_\textbf{pk}(m_1),\textit{Enc}_\textbf{pk}(m_2)\big)\Big) = m_1+m_2$.
    \item \textsl{A multiplicative operator} \textit{Mult}$(\textbf{ct}_{1},\textbf{ct}_{2})$: similar to \textit{Add} but for ciphertexts $\textbf{ct}_{1}$ and $\textbf{ct}_{2}$ of messages $m_{1}$ and $m_{2}$ and output ciphertext $\textbf{ct}_{mult}$ of $m_{1} \cdot m_{2}$. 
\end{itemize} 
To avoid to complicate the notation we will use classical symbols to denote addition and multiplication between ciphertexts.
{\color{violet} Choosing $D$ as small as possible is essential, as it is the major bottleneck for performance, in particular at the keys generation step. This
cost comes from the fact that the dimension of a ciphertext $N$ needs to grow with $D$ for a given security level $\kappa$: namely $N\geq \Omega(\kappa D)$ (refer to App.~\ref{app:ckks} for more details).}
%
{\color{violet} In this paper, we choose to use the CKKS scheme~\citep{cheon2017homomorphic} because
it supports operations on real numbers. 
}

\textbf{Other HE schemes.}
Most HE schemes~\citep{elgamal1985public, paillier1999public, rivest1978method} are
\textit{Partially Homomorphic} and only support either additions or multiplications,
but not both.
Other schemes that support any number of operations are called
\textit{Fully Homomorphic} encryption (FHE) schemes. Most LFHE schemes can
be turned into FHE schemes thanks to the bootstrapping technique introduced
by~\cite{gentry2009fully}. However, the computational cost is extremely high.
It is thus important to optimize the design of the algorithm to minimize
its multiplicative depth and (possibly) avoid bootstrapping~\citep{acar2018asurvey, ducas2015fhew, Zhao2018GeneralizedBT}.


\begin{figure}[t]
    \begin{minipage}{0.45\linewidth}
        \begin{algorithm}[H]
            \small
            \caption{Encrypted Contextual Bandit (\textcolor{cadmiumred}{Server-Side})}
            \label{alg:secure_protocol}
            \begin{algorithmic}
                \STATE {\bfseries Input:} Agent: $\mathfrak{A}$, public key: $\textbf{pk}$, horizon: $T$
                \FOR{$t=1, \ldots, T$}
                        \STATE Agent $\mathfrak{A}$ observes encrypted context $(x_{t,a})_{a\in[K]} = (\textit{Enc}_\textbf{pk}(s_{t,a}))_{a\in[K]}$
                        \STATE Agent $\mathfrak{A}$ computes the next action as a function of the encrypted history and $(x_{t,a})_{a\in[K]}$
                        and outputs an encrypted action $u_t = \textit{Enc}_\textbf{pk}(a_{t})$
                        \STATE Agent $\mathfrak{A}$ observes encrypted reward $y_{t} = \textit{Enc}_\textbf{pk}(r_{t})$
                \ENDFOR
            \end{algorithmic}
        \end{algorithm}
    \end{minipage}\hfill
    \begin{minipage}{0.45\linewidth}
        \begin{algorithm}[H]
            \small
            \caption{Encrypted Contextual Bandit (\textcolor{cadmiumred}{User-Side})}
            \label{alg:secure_protocol_user}
            \begin{algorithmic}
                \STATE {\bfseries Input:} public key: $\textbf{pk}$, secret key: $\textbf{sk}$
                \FOR{$t=1, \ldots, T$}
                        \STATE User $t$ observes features $(s_{t,a})_{a\leq K}$ and sends $(x_{t,a})_{a\in[K]} = (\textit{Enc}_\textbf{pk}(s_{t,a}))_{a\in[K]}$ to the server
                        \STATE User $t$ receives encrypted action $u_t$
                        \STATE User $t$ decrypts action $a_t = \textit{Dec}_\textbf{sk}(u_{t})$
                        \STATE User $t$ observes reward $r_{t} = r(s_{t,a_t}) + \eta_{t}$ and sends $\textit{Enc}_\textbf{pk}(r_{t})$ to the server
                \ENDFOR
            \end{algorithmic}
        \end{algorithm}
    \end{minipage}
\end{figure}
\section{CONTEXTUAL BANDIT AND ENCRYPTION}\label{sec:cbandsecurity}
A contextual bandit problem is a sequential decision-making problem with $K \in \mathbb{N}_+$ arms and horizon $T \in \mathbb{N}_+$~\citep[\eg][]{lattimore2020bandit}.
At each time $t \in [T]:=\{1,\ldots,T\}$, a learner first observes a set of features $(s_{t,a})_{a\in [K]} \subset \mathbb{R}^{d}$, selects an action $a_{t}\in [K]$ and finally observes a reward $r_t = r(s_{t,a_t}) + \eta_t$ where $\eta_t$ is a conditionally independent zero-mean noise. We do not assume anything on the distribution of the features $(s_{t,a})_{a}$.
The performance of the learner $\mathfrak{A}$ over $T$ steps is measured by the regret, that measures the cumulative difference between playing the optimal action and the action selected by the algorithm. Formally, let $a_{t}^{\star} = \arg\max_{a\in [K]} r(s_{t,a})$ be the optimal action at step $t$, then the pseudo-regret is defined as:
\begin{equation}\label{eq:regret_def}
R_{T} = \sum_{t=1}^{T}
    r(s_{t,a_t^\star}) - r(s_{t,a_t}).
\end{equation}

To protect privacy and avoid data tempering, we introduce end-to-end encryption to this protocol. Contexts and rewards are encrypted before being observed by the learner; we call this setting encrypted contextual bandit (Alg.~\ref{alg:secure_protocol}).
Formally, at  time  $t \in [T]$, the learner $\mathfrak{A}$ observes encrypted features $x_{t,a} = \textit{Enc}_{\textbf{pk}}(s_{t,a})$ for all actions $a \in \mathcal{A}$, and the encrypted reward $y_t = \textit{Enc}_{\textbf{pk}}(r_t)$
associated to the selected action $a_t$. The learner may know the public key \textbf{pk} but not the secure key \textbf{sk}. The learner is thus not able to decrypt messages and it \emph{never} observes the true contexts and rewards. {\color{violet}We further assume that both the agent $\mathfrak{A}$ and the users follow the honest-but-curious model, that is to
say each parties follow their protocol honestly but try to learn as much as possible about the other parties private data.}
\footnote{A trusted third party can be used to generate a public and secret keys. Those keys are then sent to the users but \textbf{not} to the agent $\mathfrak{A}$ (see Sec.~\ref{sec:discussion}).}
As a consequence, the learner can only do computation on the encrypted information. As a result, all the internal statistics used by the bandit algorithm are now encrypted. On user's side (see Alg.~\ref{alg:secure_protocol_user}), upon receiving an encrypted action $u_t = {Enc}_{\textbf{pk}}(a_t)$ and decrypting it $a_t = {Dec}_{\textbf{sk}}(u_t)$ using the secure key \textbf{sk}, the user generates a reward $r_t = r(s_{t,a_t}) +\eta_t$ and sends to the learner the associated ciphertext $y_t$. The learning algorithm is able to encrypt the action since the public key is publicly available. See App.~\ref{app:protocol} for additional details.


We focus on the well-known linear setting where rewards are linearly representable in the features. Formally, for any feature vector $s_{t,a}$, the reward is $r(s_{t,a}) = \langle s_{t,a}, \theta^\star \rangle$, where $\theta^\star \in \mathbb{R}^d$ is unknown. For the analysis, we rely on the following standard assumption:
\begin{assumption}\label{assumption:boundness}
    There exists $S>0$ such that $\| \theta^{\star}\|_{2}\leq S$ and there exists $L\geq 1$ such that, for all time $t \in [T]$ and arm $a \in [K]$,
    $\| s_{t,a}\|_{2}\leq L$ and $r_{t} = \langle s_{t,a}, \theta^\star \rangle + \eta_t \in [-1,1]$ {\color{violet} with $\eta_{t}$ being $\sigma$-subGaussian for some $\sigma>0$ }.
\end{assumption}
\section{AN ALGORITHM FOR ENCRYPTED LINEAR CONTEXTUAL BANDITS}\label{sec:algo}
In the previous section, we have introduced a generic framework for contextual bandit with encrypted information. Here, we provide the first algorithm able to learn with encrypted observations.
%
%
\begin{figure}[ht]
    \begin{algorithm}[H]
        \small
        \caption{Simplified \algo}
        \label{alg:simple.algo}
        \begin{algorithmic}
            \FOR{$t=1, \ldots, T$}
                    \IF{\enc{Update} ({\footnotesize Step \ding{185}}) }
                    \STATE {\footnotesize \underline{Step} \ding{182}:}
                    Estimate \enc{encrypted parameter } using $\{x_{l,a_l}, y_l\}_{l\in[t-1]}$

                    \ENDIF
                    \STATE Observe \enc{encrypted contexts} $(\enc{x_{t,a}})_{a\in[K]} = (\textit{Enc}_\textbf{pk}(s_{t,a}))_{a\in[K]}$
                    \STATE {\footnotesize \underline{Step} \ding{183}:}
                    Compute \enc{encrypted indexes} $(\enc{\rho_a(t)})_{a\in[K]}$
                    \STATE {\footnotesize \underline{Step} \ding{184}:}
                    Compute $\enc{\arg\max_a \{\rho_a(t)\}}$
            \ENDFOR
        \end{algorithmic}
    \end{algorithm}
\end{figure}
In the non-secure protocol, algorithms based on the optimism-in-the-face-of-uncertainty (OFU) principle such as \linucb~\citep{ChuLRS11} and \oful~\citep{abbasi2011improved} have been proved to achieve the regret bound $O\big( Sd\sqrt{T}\ln(TL) \big)$. Clearly, they will fail to be used as is in the secure protocol and need to be rethinked around the limitations of HE (mainly approximations in most operations).
As mentioned in the introduction, there are many, both theoretical and practical, challenges to leverage HE in this setting. Indeed, \textbf{1)} computing an estimate of the parameter $\theta^\star$ from ridge regression is extremely difficult with HE as finding the inverse of a matrix is not directly feasible
for a leveled scheme~\citep{esperancca2017encrypted}. \textbf{2)} Similarly,
computing the bonus for the optimistic action selection requires invoking operations that are not
{\color{violet} naturally} available in HE {\color{violet} hence incurring a large computational cost}. Finally, \textbf{3)} computing the maximum element (or maximum index) of a list of encrypted values is non-trivial for the algorithm alone, as it cannot observe the values to compare. In this section, we will provide HE compatible operations addressing these three issues.
{\color{violet}Each step is highly non-trivial and correctly combining them is even more challenging due to error compounding. We believe the solution we provide for each individual step may be of independent interest.}



Alg.~\ref{alg:simple.algo} report a simplified version of our HE bandit algorithm.
Informally, at each round $t$, our algorithm \algo (\emph{Homomorphically Encrypted Linear Bandits}) builds an HE estimate $\omega_t$
of the unknown $\theta^\star$ ($\omega^\star = \textit{Enc}_\textbf{pk}(\theta^\star)$) using
the observed encrypted samples, compute HE optimistic indexes $(\rho_a(t))_a$ for each action and select the action maximizing the index.
We stress that all the mentioned statistics ($\omega_t$ and $\rho_a(t)$) are encrypted values.
Indeed, \algo operates directly in the encrypted space, \ie the space
 of \emph{complex polynomials} of degree $N$.
Let's analyze those three steps.

\textbf{Step \ding{182}: HE Friendly Ridge Regression}\\[.05in]
The first step is to build an estimate of the parameter $\theta^\star$.
In the non-encrypted case, we can simply use $\theta_t = V_t^{-1} \sum_{l=1}^{t-1} s_{l,a_l} r_l$, where $V_t=\sum_{l=1}^{t-1} s_{l,a_l}s_{l,a_l}^T + \lambda I$. With
encrypted values $(x_{l,a_l}, y_l)_{l\in[t-1]}$, it is  possible to compute an encrypted matrix
$\Lambda_t = \sum_{l=1}^{t-1} x_{l,a_l} x_{l,a_l}^T+ \lambda \textit{Enc}_{\textrm{pk}}(I) =
\textit{Enc}_{\textrm{pk}}(V_t)$ and vector $\sum_{l=1}^{t-1}x_{l,a_l} y_l$ as
these operations (summing and multiplying) are HE compatible. The issue resides in the
computation of $\Lambda_t^{-1}$.  An approximate inversion scheme can be leveraged though.

Given a  matrix $V\in \mathbb{R}^{d}$ with eigenvalues $\lambda_{1} \geq \hdots \geq  \lambda_{d}>0$ and $c\in\mathbb{R}$ such that for all $i\in [d]$, $\lambda_{i}\in \text{Conv}\left(\{z\in \mathbb{R}\mid |z - c| 
\leq c \}, 2c \right)\setminus\{0, 2c\}$\footnote{$\text{Conv}(E)$ is the convex hull of set $E$.},
we define the following sequence of matrices~\citep{guo2006schur}
\begin{equation}\label{eq:iterate_inverse}
    X_{k+1} = X_{k}(2I_{d} - M_{k}),~~~ M_{k+1} = (2I_{d} - M_{k})M_{k},
\end{equation}
initialized at $X_{0} = \frac{1}{c}I_{d}$ and $M_{0} = \frac{1}{c}V$. We can show that this sequence converges to $V^{-1}$.
\begin{proposition}\label{prop:convergence_speed_approx_inv} If
  $V\in \mathbb{R}^{d\times d}$ is a symmetric positive definite matrix,
$c \geq  \text{Tr}(V)$ and for some precision level $\varepsilon>0$,
the iterate in~\eqref{eq:iterate_inverse} satisfies
$
\| X_{k} - V^{-1} \| \leq \varepsilon
$ for any $k\geq k_{1}(\varepsilon)$ with $k_{1}(\varepsilon) =
\frac{1}{\ln(2)}\ln\left( \frac{\ln(\lambda) + \ln(\varepsilon)}{\ln\left(1 -
\frac{\lambda}{c} \right)}\right)$,
where $\lambda \leq \lambda_{d}$ is a lower bound to the minimal eigenvalue of $V$
and $\| \cdot \|$ is the matrix spectral-norm.
\end{proposition}

Since $V_t$ is a regularized matrix, it holds that $\lambda_{d} \geq \lambda >0$ and by setting $c = \lambda d + L^{2}t$ we get  that $c \geq \text{Tr}(V_{t}) \geq \max_{i} \{\lambda_{i}\}$, for any step $t \in [T]$.
Therefore, we can apply iterations~\eqref{eq:iterate_inverse} to $\Lambda_t = \textit{Enc}_{\textrm{pk}}(V_t)$ since are all HE compatible operations (additions and matrix multiplications).
For $\varepsilon_t>0$, iterations~\eqref{eq:iterate_inverse} gives a $\varepsilon_t$-approximation $A_t:= X_{k_1(\varepsilon_t)}$ of $V_t^{-1}$, \ie
$\|\textit{Dec}_{\textrm{sk}}(A_t) - V_t^{-1}\| \leq \varepsilon_t$.
As a consequence,  an encrypted estimate of the unknown parameter $\theta^\star$ can be computed by mere simple matrix multiplications $\omega_t = A_t \sum_{l=1}^{t-1} {x}_{l,a_l} {y}_l$. Leveraging the concentration of the inverse matrix, the following error bound for the estimated parameter  holds.
\begin{corollary}\label{cor:distance_approx_ols}
    Setting $\varepsilon_t = \Big(Lt^{3/2}\sqrt{L^{2}t + \lambda}\Big)^{-1}$ in Prop.~\ref{prop:convergence_speed_approx_inv},
    then $\| \textit{Dec}_{\textbf{sk}}(\omega_{t}) - \theta_t\|_{V_{t}} \leq t^{-1/2}$, $\forall t$.
\end{corollary}
This result, along with the standard concentration for linear bandit~\citep[Thm.\ 2]{abbasi2011improved}, implies that, at all time steps $t$,  with probability at least $1-\delta$:
\begin{align}\label{eq:ci.inflated}
    \theta^{\star} \in \wt{\mathcal{C}}_{t} := \{ \theta\in \mathbb{R}^{d} \mid \| \textit{Dec}_{\textbf{sk}}(\omega_{t}) - \theta\|_{V_{t}} \leq \wt{\beta}_{t}\},
\end{align}
where  $\|a\|_B = \sqrt{a^\top Ba}$ and $\wt{\beta}_{t} = t^{-1/2} + S\sqrt{\lambda} + \sigma\sqrt{d\left(\ln\left(1 + L^{2}t/\lambda\right) + \ln(\pi^{2}t^{2}/(6\delta))\right)}$
is the inflated confidence interval due to the approximate inverse (see Prop.~\ref{prop:confidence_theta} in App.~\ref{app:confidence_theta}). Note that $\wt{\beta}_{t}$ is a plain scalar, not an encrypted value.

\textbf{Step \ding{183}: Computing The Optimistic Index}\\[.05in]
Once solved the encrypted ridge regression, the next step for \algo is to compute an optimistic index $\rho_a(t)$ such that $r(s_{t,a}) \lessapprox \textit{Dec}_{\textbf{sk}}(\rho_a(t))$.
For any feature vector $s_{t,a}$, by leveraging the confidence interval in~\eqref{eq:ci.inflated}, the optimistic (unencrypted) index is given by $\max_{\theta\in\wt{\mathcal{C}}_{t}} \langle \theta, s_{t,a}\rangle = \langle \textit{Dec}_{\textbf{sk}}(\omega_{t}), s_{t,a}\rangle + \wt{\beta}_{t}\|s_{t,a}\|_{V_{t}^{-1}}$.
Leveraging Prop.~\ref{prop:convergence_speed_approx_inv}, the definition of $\varepsilon_t$ in Cor.~\ref{cor:distance_approx_ols} and $\|s_{t,a}\|_2 \leq L$, it holds that:
{\small\begin{equation*}\label{eq:error_norm_inverse}
\begin{aligned}
\forall s_{t,a}, ~\|s_{t,a}\|_{V_{t}^{-1}}^{2} - \|s_{t,a}\|_{\textit{Dec}_{\textbf{sk}}(A_{t})}^{2} \leq L^{2}\|V_{t}^{-1} - \textit{Dec}_{\textbf{sk}}(A_{t})\| \\
\leq Lt^{-\frac{3}{2}}(\lambda + L^{2}t)^{-1/2}
\end{aligned}
\end{equation*}}
which leads to $\max_{\theta\in\wt{\mathcal{C}}_{t}} \langle \theta, s_{t,a}\rangle \leq \langle \textit{Dec}_{\textbf{sk}}(\omega_{t}), s_{t,a}\rangle + \sqrt{\| s_{t,a}\|_{\textit{Dec}_{\textbf{sk}}(A_{t})}^{2} + L\big(t^{3/2}\sqrt{\lambda + L^{2}t}\big)^{-1}}$. As a consequence, we can write that the encrypted optimistic index is given by:
{\small\begin{equation}\label{eq:bonus.enc.approx}
    \begin{aligned}
    &\rho_a(t) \approx \langle \omega_{t}, x_{t,a}\rangle+ \wt{\beta}_t \times\\
    \; &\times\mathrm{sqrt}_{\mathrm{HE}}\Big(
    \underbrace{x_{t,a}^\top A_{t} x_{t,a} + L\big(t^{3/2}\sqrt{\lambda + L^{2}t}\big)^{-1}}_{\text{\ding{67}}}
    \Big)
\end{aligned}
\end{equation}}
where $\mathrm{sqrt}_{\mathrm{HE}}$ is an approximate root operator in the encryption space. Unfortunately, computing the root is a non-native operation in HE and we need to build an approximation of it.

For a real value $z \in [0,1]$, we define the following sequences~\citep{cheon2019efficient}
\begin{equation}\label{eq:iterate_sqrt}
    q_{k+1} = q_{k}\left(1 - \frac{v_{k}}{2}\right),~~~ v_{k+1} = v_{k}^{2}\left(\frac{v_{k} -3}{4}\right)
\end{equation}
where $q_{0} = z$ and $v_{0} = z-1$. It is possible to show that this sequence converges to $\sqrt{z}$.
\begin{proposition}\label{prop:iterations_square_root}
For any $z \in \mathbb{R}_{+}$, $c_{1},c_{2}>0$ with $c_{2} \geq z\geq c_{1}$
and a precision $\varepsilon>0$, let $q_{k}$ be the result of $k$ iterations
of Eq.~\eqref{eq:iterate_sqrt}, with $q_{0} = \frac{z}{c_{2}}$
and $v_{0} = \frac{z}{c_{2}} - 1$.
Then, $|q_{k}\sqrt{c_{2}} - \sqrt{z}| \leq \varepsilon$ for any
$k\geq k_{0}(\varepsilon) := \frac{1}{\ln(2)}\left(\ln\left(\ln\left(\varepsilon\right)-
    \ln\left(\sqrt{c_{2}}\right) \right)- \ln\left(4\ln\left(1 - \frac{c_{1}}{4c_{2}}\right)\right)\right)$.
\end{proposition}
Therefore, by setting $z = \|x_{t,a}\|_{A_t}^2 + c_1$ (\ie as \ding{67} in Eq.~\eqref{eq:bonus.enc.approx}), $c_{1} = L(t^{3/2}\sqrt{\lambda + L^{2}t})^{-1}$, $c_{2} = c_{1} + L^{2}\lambda^{-1/2}\left( 1 +\lambda^{-1/2}\right)$ and $\varepsilon = t^{-1}$, we set
\begin{align}\label{eq:bonus.enc}
    \rho_a(t) = \langle \omega_{t}, x_{t,a}\rangle +
    \wt{\beta}_t \left(\sqrt{c_{2}}q_{k_{0}(1/t)} + \frac{1}{t}\right),
\end{align}
which implies that $r(s_{t,a}) \lessapprox \max_{\theta\in\wt{\mathcal{C}}_{t}} \langle \theta, s_{t,a}\rangle \leq \textit{Dec}_{\textbf{sk}}(\rho_a(t))$. Note that while $\omega_{t}$, $x_{t,a}$ and $q_i$ are encrypted values, $\wt{\beta}_t$, $c_1$, $c_2$ and $t$ are plain scalars.

\textbf{Step \ding{184}: HE Approximate Argmax}\\[.05in]
The last challenge faced by the learning algorithm is to compute $\arg\max_{a \in [K]} \{\rho_{a}(t)\}$. Although, it is
theoretically possible to compute an argmax procedure operating on encrypted numbers \citep{gentry2009fully}, it is highly non practical because it relies on
bootstrapping.
Recently, \citet{cheon2019efficient} introduced an homomorphic compatible algorithm (\ie approximate), called $\textbf{NewComp}$,
that builds a polynomial approximation of
$\text{Comp}(a,b)= \mathds{1}_{\{a>b\}}$ for any $a,b\in [0,1]$. This algorithm allows to compute an HE friendly approximation of $\max\{a, b\}$ for any $a,b\in [0,1]$.
We leverage this idea to derive $\textbf{acomp}$, a homomorphic compatible algorithm to compute an approximation of the maximum index (see Alg.~\ref{alg:newton_comp} in App.~\ref{app:computing_argmax}).
Precisely, $\textbf{acomp}$ does not compute $\arg\max_{a\in [K]}\{ \rho_{a}(t)\}$ but an approximate vector $b_{t} \approxeq (\mathds{1}_{\{a = \arg\max_{i} \rho_{i}(t)\}})_{a\in [K]}$.
The maximum index is the value $a$ such that $(b_t)_a$ is greater than a threshold accounting for the approximation error.

The $\textbf{acomp}$ algorithm works in two phases. First, $\textbf{acomp}$ computes an approximation $M$ of $\max_{i \in [K]} \{\rho_{i}(t)\}$ by comparing each pair $(\rho_{i}(t),\rho_{j}(t))$ with $i<j\leq K$.
Second, each value $\rho_a(t)$ is compared to this approximated maximum value $M$ to obtain $(b_t)_{a}$, an approximate computation
of $\mathds{1}_{\{\rho_a(t) > M\}}$. Cor.~\ref{cor:precision_argmax} shows that if a component of $b_{t}$ is big enough, the difference between $\max_{a} \rho_{a}(t)$ and any arm with $4(b_{t})_{a} \geq t^{-1}$ is bounded by $\wt{O}(1/t)$ (proof in App.~\ref{app:computing_argmax}).
\begin{corollary}\label{cor:precision_argmax}
    At any time $t \in [T]$,  any arm $a \in [K]$ satisfying $(b_{t})_{a} \geq \frac{1}{4t}$ is such that:
    {\small\begin{equation}
        \begin{aligned}
        &\rho_{a}(t) \geq \max_{a'\in [K]} \{\rho_{a'}(t)\} -\frac{1}{t} 
       \\&-\frac{\wt{\beta}_t}{t}\Bigg[\frac{2}{t} +  \sqrt{\frac{L}{t^{3/2}\sqrt{\lambda + L^{2}t}}} + L\sqrt{\frac{1}{\lambda} + \frac{1}{\sqrt{\lambda}}}\Bigg]
        \end{aligned}
    \end{equation}}
\end{corollary}
{\color{violet} Cor.~\ref{cor:precision_argmax} shows that while an action $a$ such that $4t(b_{t})_{a}\geq 1$ may not belong to $\arg\max_{a\in[K]} \{ \rho_{t}(a)\}$, it can be arbitrarily close, hence limiting the impact on the regret.}
As shown later, this  has little impact on the final regret of the algorithm as the approximation error decreases fast enough.
Since $b_{t}$ is encrypted, the algorithm does not know the action to play. $b_t$ is sent to the user who decrypts it and selects the action to play (the user is the only one having access to \textbf{sk}).
{\color{violet}$b_{t} \approx (\mathds{1}_{\{a = \max_{i\in [K]} \rho_{i}(t)\}}$ indicates to the user which action to take which is necessary by design of the bandit problem. However, if the user is able to invert the
polynomial functions used to compute $b_{t}$ thanks to the rescaling of the estimates $(\rho_{a}(t))_{a}$ the latter can only learn a relative ranking
for this particular user and not the actual estimates.}

\textbf{Step \ding{185}: Update Schedule}\\[.05in]
Thanks to these steps, we can prove (see App.~\ref{app:proof_inefficient_linucb}) a $\sqrt{T}$ regret bound for \algo when $\omega_t$ is recomputed at each step $t$. However, this approach would be impractical due to the extremely high number of multiplications performed.
In fact, inverting the design matrix at each step incurs a large multiplicative depth and computational cost.
The most natural way of reducing this cost is to reduce the number of times the ridge regression is solved.
The arm selection policy will not be updated at each time step but rather only when necessary.
Reducing the number of policy changes is exactly the aim of low switching algorithms~\citep[see e.g.,][]{abbasi2011improved,perchet2016batched,bai2019provably,calandriello2020near,dong2020multinomial}.
We focus on a dynamic, data-dependent batching since $\sqrt{T}$ regret is not attainable using a fixed
known-ahead-of-time schedule~\citep{han2020sequential}.

\citet{abbasi2011improved} introduced a low switching variant of \oful (\rsoful) that recomputes the ridge regression only when the following condition: $\text{det}(V_{t+1}) \geq (1 + C)\text{det}(V)$ is met, with $V$ the design matrix after the last update. The regret of \rsoful scales as $\wt{\mathcal{O}}( d\sqrt{(1 + C)T})$.
In the secure setting, computing the determinant of an encrypted matrix is costly~\citep[see e.g.][]{kaltofen2005complexity} and requires multiple matrix multiplications. The complexity of checking the above condition with HE outweights the benefits introduced by the low switching regime, rendering this technique  non practical. Instead of a determinant-based condition, we consider a trace-based condition, inspired by the update rule for GP-BUCB~\citep{desautels2014parallelizing, calandriello2020near}.

The ``batch $j$'' is defined as the set of time steps between $j$-th and $(j+1)$-th updates of $\omega$, and we
denote by $t_{j}$ the first time step of this batch.
The design matrix is now denoted by
$\wb{\Lambda}_j = \lambda \textit{Enc}_{\textbf{pk}}(I) + \sum_{l=1}^{t_j-1} {x}_{l,a_l} {x}_{l,a_l}^{\intercal}$,
and more importantly is only updated at the beginning of each batch $j$ (and similarly for the inverse $\wb{A}_{j}$ and vector $\omega_{j}$).
The current batch $j$ is ended if and only if the following \emph{trace-based condition} is met at some time $t$:
\begin{align}\label{eq:trace_condition}
    C\leq \text{Tr}\Bigg(\sum_{l = t_{j}+1}^{t-1}\wb{A}_j x_{l,a_l}x_{l,a_l}^{\intercal}\Bigg) = \sum_{l=t_{j}+1}^{t-1} \| {x}_{l,a_l} \|_{\wb{A}_{j}}^{2}
\end{align}
The intuition behind this condition is that the trace of $\wb{V}_{j} = \lambda I + \sum_{l=1}^{t_j-1} {s}_{l,a_l} {s}_{l,a_l}^{\intercal}$ is enough to directly control the regret. The following proposition shows that the error due to the computation in the encrypted space remains small.
\begin{proposition}
    Let $\varepsilon_j = \Big(Lt_{j}^{3/2}\sqrt{\lambda + L^{2}t_{j}} \Big)^{-1}$ and $\wb{A}_j = X_{k_1(\varepsilon_j)}$ as
    in Eq.~\eqref{eq:iterate_inverse} starting from $M_0= \wb{\Lambda}_j/c$ with $c\geq \lambda + t_{j}L^{2}$. Then, for any $j>0$:
        $\Big| \text{Tr}\Big(\sum_{l = t_{j}+1}^{t-1} \Big(\textit{Dec}_{\textbf{sk}}(A_{j}) - \wb{V}_{j}^{-1}\Big) s_{l,a_l}s_{l,a_l}^{\intercal}\Big) \Big| \leq L^{2}\varepsilon_{j}(t-1 -t_{j})$.
\end{proposition}
Since the switching condition involves data-dependent encrypted quantities, we leverage {\color{violet} a similar procedure as to compare indexes}.
We compute an (\emph{encrypted}) homomorphic approximation of the sign {\color{violet} function thanks to the \textbf{acomp} algorithm. The result is an encryption
of the approximation of $\mathds{1}_{\{\}}$. Similarly to computing the argmax of $(\rho_{a}(t))_{a}$, the algorithm cannot access the result, thus it relies on the user to decrypt and send the result of the comparison to decide
whenever the algorithm needs to update the approximate inverse $\bar{A}_{j}$, . However,
to prevent any information leakage, that is to say the algorithm or the user learning about the features of other users, we use a masking procedure which obsfucates the result of the decryption to the user (detailed in App.~\ref{app:trace_condition} and App.~\ref{app:masking_procedure}).}


In non-encrypted setting, Cond.~\ref{eq:trace_condition} can be used to dynamically control the growth of the regret, that is bounded by
$\mathcal{O}\big(\sum_{j=0}^{M_T} \sum_{t = t_j +1}^{t_{j+1}} \|\wb{V}_{j}^{-1/2}s_{t,a_{t}}\|_{2}\big)$. But in the
secure setting, the regret can not be solely bounded as before. The condition for updating the batch has to take into account
the approximation error introduced by all the approximate operations.            %
Let $M_T$ be the total number of batches, then the contribution of the approximations to the regret scales as $\sum_{j=0}^{M_{T}-1} \wt{\mathcal{O}}((t_{j+1} - t_{j})^2\varepsilon_{j})$.
We thus introduce an additional condition aiming at explicitly controlling the length of each batch.
Let $\eta>0$, then a new batch is started if Cond.~\eqref{eq:trace_condition} is met or if: $t\geq (1 + \eta)t_{j}$. 
This ensures that the additional regret term grows proportionally to the total number of batches $M_{T}$.
Note that $t_j$ and $t$ are not encrypted values and the comparison is ``simple''. The full algorithm is reported in App.~\ref{app:full_algo}.

\section{THEORETICAL GUARANTEES}\label{sec:regret}
The regret analysis of \algo is decomposed in two parts. First, we show that, the number of batches is logarithmic in $T$. Then, we bound the error of approximations per batch.
\begin{proposition}\label{prop:nb_batches}
    For any $T>1$, if $ C - \frac{L\eta}{\sqrt{\lambda + L^{2}}} > \frac{1}{4}$, the number of episodes $M_T$ of \algo (see Alg.~\ref{alg:simple.algo}) is bounded by:
    \begin{align}
        M_{T} \leq 1 + \frac{d\ln\left(1 + \frac{L^{2}T}{\lambda d}\right)}{2\ln\left( \frac{3}{4} + C - \frac{L\eta}{\sqrt{\lambda + L^{2}}} \right)} + \frac{\ln(T)}{\ln(1 + \eta)}
    \end{align}
\end{proposition}
The total number of multiplications to compute $\omega_j$ is $T/M_T$-times smaller thanks to the low-switching
condition. This leads to a vast improvement in computational complexity.
Note that at each round $t$, \algo still computes the upper-confidence
bound on the reward and the maximum action.
Leveraging this result, when any of the batch conditions is satisfied, the regret can be controlled in the same way as the non-batched case, up to a multiplicative constant.
\begin{theorem}\label{thm:regret_efficient_linucb}
Under Asm.~\ref{assumption:boundness}, for any $\delta>0$ and $T\geq d$, there exists constants $C_1,  C_2>0$ such that the regret of \algo (Alg.~\ref{alg:simple.algo}) is bounded with probability $1 - \delta$ by:
{\small
\begin{equation*}
    \begin{aligned}
        R_{T} \leq  C_{1}\beta^{\star}\Bigg(\sqrt{\left(1.25+ C\right)dT\ln\left(\frac{TL}{\lambda d} \right)} + \frac{L^{3/2}}{\sqrt{\lambda}}\ln(T)\Bigg)&\\
        + C_2 \beta^{\star}M_{T}\max\left\{\sqrt{L} + \frac{\eta }{\sqrt{L}}, \eta^{2} + \frac{L}{\sqrt{\lambda + L^{2}}^{3}}\right\}&
    \end{aligned}
\end{equation*}
}
with $\beta^{\star} = 1 + \sqrt{\lambda}S + \sigma\sqrt{d\left(\ln\left(1 + \frac{L^{2}T}{\lambda d}\right) +  \ln\left(\frac{\pi^{2}T^{2}}{6\delta}\right)\right)}$
and $M_T$ as in Prop.~\ref{prop:nb_batches}.
\end{theorem}
The first term of the regret highlights the impact of the approximation of the square root and maximum that are computed at each round.
The second term shows the impact of the approximation of the inverse. It depends on the number of batches since the inverse is updated only once per batch. By Prop.~\ref{prop:nb_batches}, we notice that this term has a logarithmic impact on the regret.
Finally, the last term is the regret incurred due to low-switch of the optimistic algorithm. We can notice that the parameter $C$ regulates a trade-off between regret and computational complexity. This term is also the regret incurred by running \oful with trace condition instead of the determinant-based condition. This further stress that the cost of encryption on the regret is only logarithmic, leading to a regret bound of the same order of the non-secure algorithms.
{\color{violet} But the computationnal complexity of \algo is multiple orders higher than any non-encrypted bandit algorithm. For example the complexity of computing a scalar product with HE now scales with the ring dimension $N$ and not the dimension of the contexts anymore $d \ll N$.}\todompout{Can we add a paragraph about computational complexity?}


\section{DISCUSSION AND EXTENSIONS}\label{sec:discussion}

In this section, we present a numerical validation of the proposed algorithm in a secure linear bandit problem and we discuss limitations and possible extensions.
        \begin{wrapfigure}{l}{0.3\textwidth}
            \begin{center}
             \includegraphics[width=\linewidth]{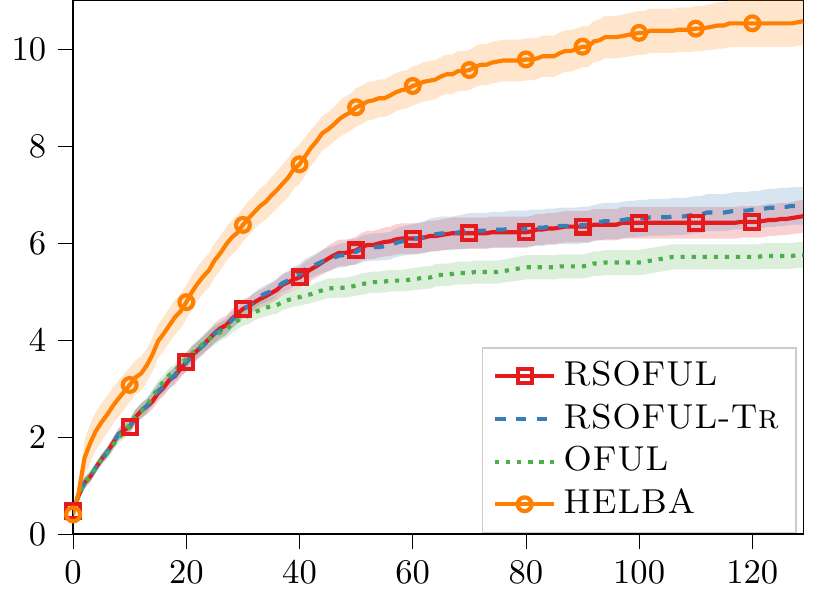}
            \end{center}
            \caption{\label{fig:regret_secofulls} Regret on a toy problem with $4$ random uniform contexts.}
          \end{wrapfigure}
    \textbf{Numerical simulation.}
        Despite the mainly theoretical focus of the paper, we illustrate the performance of the proposed algorithm on a toy example, where
        we aim at empirically validating the theoretical findings.
        We consider a linear contextual bandit problem with $4$ contexts in dimension $2$ and $2$ arms.
        As baselines, we consider \oful, \rsoful and \rsofultr (a version
        of \rsoful where  the determinant-based condition is replaced by the trace-condition in~\eqref{eq:trace_condition}).
        We run these baselines on non-encrypted data and compare the performance with \secofulls working with encrypted data.
        In the latter case, at each step, contexts and rewards are encrypted using the CKKS~\citep{cheon2017homomorphic} scheme with parameter $\kappa = 128$, 
        $D = 100$ and $N= 2^{16}$, a modulus $\log(q_{0}) = 4982$ and a cyclotomic degree of $M = 131072$ {\color{violet} chosen automatically by the PALISADE library \citep{PALISADE} used for the implementation}. {\color{violet} The size of the ciphertext is not allowed to grow and a relinearization
        is performed after every operation. The variance of the noise} in the
        reward is $\sigma = 0.5$. Finally, we use $C = 1$ and $\eta = 0.1$ in \algo. The regularization parameter is set to $1$ and $L = 5.5$.
        Fig.~\ref{fig:regret_secofulls} shows the regret of the algorithms averaged over $25$ repetitions.
        We notice that while the non-encrypted low-switching algorithms (i.e., \rsoful and \rsofultr) recompute the ridge regression only 11 times on average, their performance is only slightly affected by this and it is comparable to the one of \oful.
        The reduced number of updates is a significant improvement in light of the current limitation in the multiplicative depth of homomorphic schemes. This was the enabling factor to implement \secofulls.
        Note that the update condition in \secofulls 
         increases the number of updates to about $20$ on average.
        As expected, the successive approximations and low-switching combined worsen the regret of \secofulls. However, this small loss in performance comes with a provable guarantee on the security of users' data. 

   \textbf{Computational Complexity.}
        Even though we reduced the number of multiplications and additions, the total runtime of
        \secofulls is still significant, several orders of magnitude higher compared to the unencrypted setting, {\color{violet} the total time for $T=130$ steps
        and $\kappa = 128$ bits was $20$ hours and $39$ minutes}. We believe that a speed up can be obtained by optimizing how matrix multiplication is handled.
        For example, implementation optimization can increase the speed of computation
        of logistic regression \citep{blatt2020secure}. 
        However, we stress that \secofulls is {\color{violet} almost (up to the masking procedure)} agnostic to the homomorphic scheme used, hence any improvement in the HE literature can be leveraged by our algorithm.
        Bootstrapping procedures~\citep{gentry2009fully} can be used for converting a leveled schema into {a \textit{Fully HE} scheme}.
        This mechanism, together with the
        low-switching nature of our algorithm, can be the enabling tool for
        scaling this approach to large problems {\color{violet} as the multiplicative depth scales linearly with the dimension}.

        \textbf{Discussion.} Many other approaches are possible to increase the computational efficiency,
        for example using a trusted execution environment~\citep{sabt2015trusted} or leveraging user-side computational capacities.
        We decided to design an algorithm where the major computation (except for comparisons) are done server-side, having in mind cloud-computing or recommendations running on mobile phone.
        The objective was to make as secure as possible this protocol so that the server can leverage the information coming from all users.
        However,
        if we assume that users have greater computation capabilities, the
        algorithm can delegate some computations~\citep[see e.g.,][]{blatt2020secure}.
        For example, for the inverse, the algorithm can generate a random (invertible) matrix $N_{t}$, homomorphically compute
        $V_{t}N_{t}$ and sends the masked matrix, $V_{t}N_{t}$ to the user. The latter decrypts, inverts, re-encrypts
        the inverse and sends it to the algorithm (see \citep[Sec.~$8$]{bost2015machine} for more details).
        A similar scenario, can be imagined for computing a square root or a matrix multiplication.
        This protocol requires users to perform computationally heavy operations (inverting a
        matrix) locally.
        To ensure security with this delegation, a verification step is needed~\citep[see e.g.,][]{bost2015machine}
        further increasing communications between the user and the bandit algorithm.
        {\color{black}We believe that an interesting direction for future work is to integrate this protocol in a distributed setting (i.e., federated learning).}
        Using a server-side trusted execution environment can speed up computations as operations
        are executed in the clear in private regions of the memory.

        \textbf{Multi-users Setting.}
        Usually contexts represent different
        users, described by their features $s_{t}$ {\color{violet} and some users may want to use their own
        public key $\textbf{pk}_{t}$ (and secret key $\textbf{sk}_{t}$) to encrypt those features. In that case,
        \algo can be used with a KeySwitching \cite{fan2012somewhat,brakerski2012fully,
        brakerski2014leveled} component}.
        This operation takes a ciphertext
        $c_{1}$ decipherable by a secret key $\textbf{sk}_{1}$ and output a ciphertext
        $c_{2}$ decipherable by a secret key $\textbf{sk}_{2}$. A user send the encrypted context/reward
        to the bandit algorithm which perform a key switching (see App.~\ref{app:keyswitch_protocol}) {\color{violet} with the help of trusted third party
        who generate the set of keys used by the learning algorithm} such that all ciphertexts received are decipherable by the same key
        and compatible for homomorphic operations. KeySwitching can be performed without accessing the data
        and {\color{violet}with some (or all) users using their own set of private/public keys for encryption/decryption}. 

\vspace{-0.3cm}
\section{CONCLUSION}
\vspace{-0.3cm}
{\color{violet} In this paper, we introduced the problem of encrypted linear contextual bandits and
provided an algorithm, \algo, with a regret similar to regret bounds achievable in the non-encrypted setting.
This algorithm trades-off the approximation error and computational complexity of HE and the need
for accurate estimation to obtain sublinear regret.
We leave as open question the design of an algorithm tailored to the characteristics of
the HE and extensions to either other algorithms (e.g., Thompson sampling) or settings (e.g., reinforcement learning).}

\newpage{}
\subsubsection*{Acknowledgments}

V. Perchet acknowledges support from the French National Research Agency (ANR) under grant
number \#ANR-$19$-CE$23$-$0026$ as well as the support grant, as well as from the grant "Investissements
d’Avenir"  (LabEx Ecodec/ANR-$11$-LABX-$0047$).

\bibliography{bibliography}
\bibliographystyle{plainnat}



\clearpage
\begin{appendix}
\addcontentsline{toc}{section}{Appendix} 
\part{Appendix} 
\parttoc 

\section{SLOW-SWITCHING ALGORITHM}\label{app:full_algo}

In this section, we present the detailed algorithm of Sec.~\ref{sec:algo}.
\begin{algorithm}[H]
    \small
    \caption{Low-Switching \algo ({\color{cadmiumred}} Server-Side)}
    \label{alg:low_switching_algo}
    \begin{algorithmic}
        \STATE {\bfseries Input:} horizon: $T$, regularization factor: $\lambda$, failure probability: $\delta$,
                feature bound: $L$, $\theta^{\star}$ norm bound: $S$, dimension: $d$, batch growth: $\eta$,
                trace condition: $C$
                \STATE Set $w_{1} = Enc_{\text{pk}}(0)$, $\Lambda_{1} = Enc_{\textbf{pk}}(\lambda I)$,
                $\bar{A}_{1} = Enc_{\textbf{pk}}(\lambda^{-1}I)$, $\check{V}_{1} = Enc_{\textbf{pk}}(\lambda I)$
                $\check{g}_{0} = 0$, $j = 0$ and $t_{0} = 1$\;
                \FOR{$t=1 \hdots, T$}
                    \STATE Set $\wt{\beta}(t) = \sigma\sqrt{d\ln\Big(\Big(1 + \frac{L^{2}t_{j}}{\lambda}\Big)\Big(\frac{\pi^{2}t^{2}}{6\delta}\Big)\Big)} + t_{j}^{-1/2} + S\sqrt{\lambda}$
                    and $\epsilon_{j} = L(t_{j}^{3/2}\sqrt{\lambda + L^{2}t_{j}})^{-1}$
                    \STATE Observe \enc{encrypted contexts} $(\enc{x_{t,a}})_{a\in[K]} = (\textit{Enc}_\textbf{pk}(s_{t,a}))_{a\in[K]}$
                        \FOR{$a = 1, \hdots, K$}
                            \STATE Compute \enc{approximate square root} $\mathrm{sqrt}_{\mathrm{HE}}\Big(
    x_{t,a}^\top \bar{A}_{j} x_{t,a} + \varepsilon_{j}\Big)$
                            \STATE Compute \enc{encrypted indexes} $\rho_{a}(t) = \langle x_{t,a},w_{j} \rangle + \wt{\beta}(t)\Big(\mathrm{sqrt}_{\mathrm{HE}}\Big(
                                x_{t,a}^\top \bar{A}_{j} x_{t,a} + \varepsilon_{j}\Big) + t^{-1}\Big)$ (\footnotesize Step \ding{183})
                            \STATE  Rescale \enc{encrypted indexes} $\wh{\rho}_{a}(t) = \frac{\rho_{a}(t) - r_{\min}}{\rho_{\max} - r_{\min}}$
                            with $\rho_{\max} = r_{\max} + 2\wt{\beta}(t)\Big[\frac{2}{t} +  \frac{L}{t^{3/2}\sqrt{\lambda + L^{2}t}} + L^{2}\left(\frac{1}{\lambda} + \frac{1}{\sqrt{\lambda}}\right)\Big]$
                        \ENDFOR
                        \STATE Compute \enc{comparison vector} $b_{t}\in\mathbb{R}^{K}$
                            using \textbf{acomp} (see Alg.~\ref{alg:newton_comp} in App.~\ref{app:computing_argmax}) with precision $\varepsilon_{t}' = (4.1t)^{-1}$ (\footnotesize Step \ding{184})

                        \STATE Observe \enc{encrypted reward} $y_{t}$ and \enc{encrypted context} $x_{t,a_{t}}$
                        \STATE Update $\check{V}_{t+1} = \check{V}_{t} + x_{t,a_{t}}x_{t,a_{t}}^{\intercal}$ and $\check{g}_{t+1} = \check{g}_{t} + y_{t}x_{t,a_{t}}$
                        \STATE Compute Cond.~\eqref{eq:trace_condition} by \enc{computing $\delta_t$} with $\varepsilon = 0.45$ and
                        $\varepsilon_{t}' = L^{2}(\frac{1}{\lambda} + \frac{1}{\sqrt{\lambda}})(t-1-t_{j}))$ (see App.~\ref{app:trace_condition}).
                        \STATE Use masking procedure on $\delta_{t}$ (Alg.~\ref{alg:masking_procedure}) and sends the masked ciphertext to the user
                        \IF{$\delta_{t} \geq 0.45$ \textbf{or} $t \geq (1 + \eta)t_{j}$}
                            \STATE Set $t_{j+1} = t$, $j  = j+1$ and $\Lambda_{j+1} = \check{V}_{t}$
                            \STATE \enc{Compute $\bar{A}_{j+1} = X_{k_{1}(\varepsilon_{j+1}/L^{2})}$} as in Prop.~\ref{prop:convergence_speed_approx_inv} ($V = \Lambda_{j+1}$, $c = \lambda d + L^{2}t_{j+1}$)
                            and $w_{j+1} = \bar{A}_{j+1} \check{g}_{t_{j+1}}$
                        \ENDIF
                \ENDFOR
    \end{algorithmic}
\end{algorithm}


\section{ADDITIONAL RELATED WORK}\label{app:relwork}
In Federated Learning (a.k.a., collaborative multi-agent),  DP and LDP guarantees can provide a higher level of privacy at a small regret cost, leveraging collaboration between users \citet{wang2020global,zhu2020federatedbandit}.
Another collaborative approach to privacy-preserving machine learning, called Secure Multi-Party Computation (MPC)~\citep[e.g.][]{damgard2011multiparty},  divides computations between parties,  while guarantying that it is not possible for any of them to learn anything about the others. This has been recently empirically investigated in the bandit framework \cite{hannun2019privacy}. However, there is an additional strong assumption, that each party provides a subset of the features observed at each round.

Finally, Homomorphic Encryption (HE)~\citep[e.g.][]{halevi2017homomorphic} aims at providing a set of tools to perform computation on encrypted data, outsourcing computations to potentially untrusted third parties (in our setting the bandit algorithm) since data cannot be decrypted.
HE has  only been merely used  to encrypt rewards in bandit problems \citet{ciucanu2020secure, ciucanu2019secure}, but in some  inherently simpler setting: i)  contexts are not considered and arms' features are not encrypted; ii) a trusted party decrypts data. In particular, the second point makes   algorithm design much easier but requires  users to trust the third party which, in turn, can lead again to privacy/security concerns.
{\color{violet} In the supervised learning literature, HE has been used to train neural networks \citep{badawi2020alexnet} achieving $77.55\%$ classification accuracy on CIFAR-$10$ (compared to a state-of-the-art accuracy of
$96.53\%$ \citep{graham2015fractional}) highlighting the potentially high impact of the approximation error due to HE.}


\section{PROTOCOL DETAILS}\label{app:protocol}
{\color{black}
The learning algorithm may try to break encryption by inferring a mapping between ciphertexts
and values or by storing all data. 
HE relies on the hardness of the \emph{Learning With Error} problem~\citep{albrecht2015concrete} to guarantee security. To break an HE scheme, an attacker has to perform at least $2^{\kappa}$ operations to be able to differentiate noise from messages in a given ciphertext. We refer to~\citep{HomomorphicEncryptionSecurityStandard} for a survey on the actual number of operations needed to break HE schemes with most of the known attacks. Although collecting multiple ciphertexts may speedup some attacks, the security of any HE scheme is still guaranteed as long as long the number of ciphertexts observed by an attacker is polynomial in $N$~\citep{regev2009lattices}.
}

\subsection{CKKS Encryption Scheme}\label{app:ckks}

    In this section, we introduce the CKKS scheme \cite{cheon2017homomorphic}. This scheme is inspired
    by the BGV scheme \cite{brakerski2014leveled} but has been modified to handle the encryption of real numbers.
    The security of those schemes relies on the assumption of hardness of the Learning With Errors (LWE),
    ring-LWE (RLWE) \cite{regev2009lattices}. The scheme can be divided into $2$ parts: encoding/decoding and encryption/decryption.

    \subsubsection{Encoding and Decoding of Messages.}

        In CKKS, the space of message is defined as $\mathbb{C}^{N/2}$ for some big even integer $N\in\mathbb{N}$.
        This integer is a parameter of the scheme chosen when generating the private and secret keys.
        CKKS scheme does not work directly on the space $\mathbb{C}^{N/2}$ but rather
        on an integer polynomial ring $\mathcal{R} = \mathbb{Z}[X]/\left( X^{N} + 1\right)$ (the plaintext space)
        \cite{seidenberg1978constructions}.
        Encoding a message $m\in \mathbb{C}^{N/2}$ into the plaintext space
        $\mathcal{R}$ is not as straightforward as using a classical embedding of a vector into a polynomial
        because we need the coefficients of the resulting polynomial to be integers. To solve this issue the CKKS
        scheme use a more sophisticated construction that the canonical embedding, based on the subring
        $\mathbb{H} = \{ z\in \mathbb{C}^{N} \mid z_{j} = \bar{z}_{N-j}, j\leq N/2\}$ which is isomomorphic to
        $\mathbb{C}^{N/2}$. Finally, using a canonical embedding $\sigma: \mathcal{R} \rightarrow \sigma(\mathcal{R}) \subset \mathbb{H}$
        and the \textit{coordinate-wise random rounding} technique developed in
        \cite{lyubashevsky2013toolkit}, the CKKS scheme is able to construct an isomorphism
        between $\mathbb{C}^{N/2}$ and $\mathcal{R}$.

    \subsubsection{Encryption and Decryption of Ciphertexts.}

        Most public key scheme relies on the hardness of the \textit{Learning with Error} (LWE) problem introduced
        in \cite{regev2009lattices}. The LWE problem consists in distinguishing between noisy pairs
        $(a_{i}, \langle a_{i}, s\rangle + e_{i})_{i\leq n} \subset (\mathbb{Z}/q\mathbb{Z})^{n} \times \mathbb{Z}/q\mathbb{Z}$
        and uniformly sampled pairs in $(\mathbb{Z}/q\mathbb{Z})^{n} \times \mathbb{Z}/q\mathbb{Z}$ where
        $(e_{i})_{i\leq n}$ are random noises and $q\in \mathbb{N}$. However, building a cryptographic public key system based on
        LWE is computationally inefficient. That's why CKKS relies on the \textit{Ring Learning with Error}
        (RLWE) introduced in \cite{lyubashevsky2013ideal} which is based on the same idea as LWE but working
        with polynomials $\mathbb{Z}_{q}[X]/(X^{N}+1)$ instead on integer in $\mathbb{Z}/q\mathbb{Z}$.
        RLWE (and LWE) problem are assumed to be difficult to solve and are thus used as bases for cryptographic system.
        The security of those problems can be evaluated thanks to \cite{albrecht2015concrete} which gives practical bounds on
        the number of operations needed for known attacks to solve the LWE (RLWE) problem.

        The CKKS scheme samples a random $s$ on $\mathcal{R}$ and defines the secret key as $\textbf{sk} = (1,s)$.
        It then samples a vector $a$ uniformly on $\mathcal{R}/q_{L}\mathcal{R}$ (with $q_{L} = 2^{L}q_{0}$ where $L$ is the depth of
        the scheme and $q_{0}$ its modulus) and an error term $e$ sampled on $\mathcal{R}$ (usually each coefficient
        is drawn from a discrete Gaussian distribution). The public key is then defined as $\textbf{pk} = (a, -a.s + e)$.
        Finally, to encrypt a message $m\in \mathbb{C}^{N/2}$ identified by a plaintext $\mathfrak{m} \in \mathcal{R}$
        the scheme samples an encrypting noise $\nu\sim \mathcal{ZO}(0.5)$\footnote{A random variable $X \sim\mathcal{ZO}(0.5)$ such that
        $X\in \{0, 1, -1\}^{N}$, $(X_{i})_{i\leq N}$ are i.i.d such that for all $i\leq N$ $\mathbb{P}(X_{i} = 0) = 1/2$,$\mathbb{P}(X_{i} = 1) = 1/4$
        and $\mathbb{P}(X_{i} = -1) = 1/4$}. The scheme then samples $e_{0}, e_{1} \in \mathbb{Z}^{N}$
        two independent random variable from any distribution on $\mathcal{R}$, usually a discrete Gaussian distribution. The ciphertext associated to the message
        $m$ is then $[(\nu \cdot\textbf{pk} + (\mathfrak{m} + e_{0}, e_{1}))]_{q_{L}}$ with $[.]_{q_{L}}$ the modulo operator and
        $q_{L} = 2^{L}$. Finally, to decrypt a ciphertext $c = (c_{0}, c_{1}) \in \mathcal{R}_{q_{l}}^{2}$ (with $l$ the level
        of the ciphertext, that is to say the depth of the ciphertext), the scheme computes the plaintext $\mathfrak{m}' = [c_{0} + c_{1}s]_{q_{l}}$\footnote{ for any $n\in\mathbb{N}$, $[.]_{n}$ is the remainder of the division
        by $n$}
        and returns the message $m'$ associated to the plaintext $\mathfrak{m}'$.


    \subsection{Key Switching}\label{app:keyswitch_protocol}

        Homomorphic Encryption schemes needs all ciphertexts to be encrypted
        under the same public key in order to perform additions and multiplications.
        As we mentioned in Sec.~\ref{sec:discussion} one way to circumvent this issue
        is to use a \textit{KeySwitching} operation. The \textit{KeySwitching} operation takes as input a
        cyphertext $c_{1}$ encrypted thanks to a public key $\text{pk}_{1}$ associated
        to a secret key $\text{sk}_{1}$ and transform it into a cyphertext encrypting
        the same message as $c_{1}$ but under a different secret key $\text{sk}_{2}$.

        The exact \textit{KeySwitching} procedure for each scheme is different. We will use the
        CKKS scheme,  inspired by the BGV scheme \cite{brakerski2014leveled},  where  \textit{KeySwitching} relies on two operations $\text{BitDecomp}$ and $\text{PowerOf}2$, described below,
        \begin{enumerate}
            \item $\text{BitDecomp}(c, q)$ takes as input a ciphertext $c\in\mathbb{R}^{N}$ with $m$ the size of the ring dimension
            used in CKKS and an integer $q$. This algorithm decomposes $c$ in its
            bit representation $(u_{0}, \hdots, u_{\left \lceil  \log_{2}(q) \right \rceil}) \in \mathbb{R}^{N\times\left \lceil \log_{2}(q)\right \rceil}$
            such that $c = \sum_{j=0}^{\left\lfloor \log_{2}(q)\right\rfloor} 2^{j}u_{j}$

            \item $\text{PowerOf}2(c, q)$ takes as input a ciphertext $c\in\mathbb{R}^{N}$ and an integer $q$. This algorithm
            outputs $(c, 2c, \hdots, 2^{\left\lfloor \log_{2}(q) \right\rfloor}c) \in \mathbb{R}^{m\times \left\lceil \log_{2}(q)\right\rceil}$
        \end{enumerate}
        The \textit{KeySwitching} operation can then be decomposed as:
        \begin{itemize}
            \item the first party responsible for $\text{sk}_{1}$ generates a new (bigger, in the sense that the parameter $N$ is bigger than $\text{sk}_{1}$)
            public key $\tilde{\text{pk}}_{1}$ still associated to $\text{sk}_{1}$
            \item the owner of secret key $\text{sk}_{2}$ computes $\text{PowerOf2}(\text{sk}_{2})$ and add it to $\tilde{\text{pk}}_{1}$. This object is called the \textit{KeySwitchingKey}.
            \item the new cyphertext is computed by mulitiplying $\text{BitDecomp}(c_{1})$ with the KeySwitchingKey. This gives a new cyphertext decryptable with the secret key $\text{sk}_{2}$ and encrypted under a new public key $pk_{2}$
        \end{itemize}

        \begin{algorithm}[h]
            \caption{KeySwitching Procedure}
            \label{alg:key_switch_procedure}
            \begin{algorithmic}
                \STATE {\bfseries Input:} Cyphertext: $c$, User: $u$, User public key/secret key: $\text{pk}_{u}, \text{sk}_{u}$, Bandit  Algorithm: $\mathfrak{A}$, Trusted Third Party: $\mathfrak{B}$, integer $q$
                \STATE Alg.~$\mathfrak{A}$ receives cypthertext $c$ encrpyted with key $\text{pk}_{u}$
                \STATE $\mathfrak{B}$ sends public key $\textbf{pk}$ to $u$
                \STATE $u$ computes $\text{Enc}_{\textbf{pk}_{u}}(\textbf{ksk}_{u}) = \text{Enc}_{\textbf{pk}_{u}}(\text{PowerOf}2(\textbf{sk}_{u}, q) + \textbf{pk})$
                \STATE $u$ sends $\text{Enc}_{\textbf{pk}_{u}}(\textbf{ksk}_{u})$ to $\mathfrak{A}$
                \STATE $\mathfrak{A}$ computes the new cyphertext $c' = \text{Enc}_{\textbf{pk}_{u}}(\text{BitDecomp}(c, q)^{\intercal}) \text{Enc}_{\textbf{pk}_{u}}(\textbf{ksk}_{u}) = \text{Enc}_{\textbf{pk}_{u}}(\text{Enc}_{\textbf{pk}}(c))$
                \STATE $u$ decrypts $c'$ and sends the result to $\mathfrak{A}$

            \end{algorithmic}
        \end{algorithm}

        Alg.~\ref{alg:key_switch_procedure} allows us to perform the \textit{KeySwitching} in a private manner for the CKKS scheme.
        Indeed, the key switch operation requires to decompose a secret key thanks to the $\text{PowerOf}2$ procedure. If not done in a secure
        fashion this could lead to a leak of the frist private key. It is thus necessary to ensure that
        this key is not distributed in the clear. However, our private procedure requires communication between the
        bandit algorithm $\mathfrak{A}$ and the user $u$. In particular, the user still needs to receives the public key from
        the trusted third party. However, the user does not need to be known ahead of time as previously.


\section{TOWARD AN ENCRYPTED OFUL }

    In this section, we provide the proof of the results of Step \ding{182}, \ding{183} and \ding{184}, i.e., the speed of convergence of iterating
    Eq.~\eqref{eq:iterate_inverse} or Eq.~\eqref{eq:iterate_sqrt}, how to build a confidence intervals around $\theta^{\star}$ and how the approximate
    argmax is computed in Alg.~\ref{alg:simple.algo}.

    \subsection{Computing an Approximate Inverse}\label{app:proof_inverse}

        First, we prove Prop.~\ref{prop:convergence_speed_approx_inv}. The proof of convergence the Newton method for matrix inversion is rather standard but
        the proof of convergence for the stable
        method (Eq.~\eqref{eq:iterate_inverse}) is often not stated.
         We derive it here for completeness.
         First, we recall Prop.~\ref{prop:convergence_speed_approx_inv}.

        \begin{proposition*}

            Given a symmetric positive definite matrix $V\in \mathbb{R}^{d\times d}$,
            $c \geq  \text{Tr}(V)$ and  a precision level $\varepsilon>0$,
             the iterate in~\eqref{eq:iterate_inverse} satisfies
             $$
                 \| X_{k} - V^{-1} \| \leq \varepsilon
             $$
             for any $k\geq k_{1}(\varepsilon)$ with
             $$k_{1}(\varepsilon) =
             \frac{1}{\ln(2)}\ln\left( \frac{\ln(\lambda) + \ln(\varepsilon)}{\ln\left(1 -
             \frac{\lambda}{c} \right)}\right)$$,
              where $\lambda \leq \lambda_{d}$ is a lower bound to the minimal eigenvalue of $V$
              and $\| \cdot \|$ is the matrix spectral-norm.

        \end{proposition*}

        \begin{proof}{of Prop.~\ref{prop:convergence_speed_approx_inv}.} After $k$ iterations of Eq.~\eqref{eq:iterate_inverse}, we have that $VX_{k} = M_{k}$. Indeed we proceed by induction:
                \begin{itemize}
                    \item For $k= 0$, $M_{0} = \frac{1}{c} V = VX_{0}$
                    \item For $k+1$ given the property at time $k$, $VX_{k+1} = VX_{k}(2I_{d} - M_{k}) = M_{k}(2I_{d} - M_{k}) = M_{k+1}$
                \end{itemize}
            Let's note $E_{k} = X_{k} - V^{-1}$ and $\tilde{E}_{k} = M_{k} - I_d$ then:
            \begin{align*}
                E_{k+1} =  \left(X_{k+1}V - I_{d} \right)V^{-1} &= \left(M_{k+1} - I_{d}\right)V^{-1}\\
                &= -\left( M_{k}^{2} - 2M_{k} + I_{d}\right)V^{-1}\\
                &= -\left(M_{k} - I_{d}\right)^{2}V^{-1} = -\tilde{E}_{k}^{2}V^{-1}
            \end{align*}
            where the second equality is possible because $V$ and $(X_{k})_{k\in \mathbb{N}}$ commute as for all
            $k\in \mathbb{N}$, $X_{k}$ is a polynomial function of $V$.

            Therefore, we have for any $k\in \mathbb{N}$:
            \begin{equation}
                \|E_{k+1}\| = \| \tilde{E}_{k}^{2} V^{-1}\| \leq \|V^{-1}\| \times\|\tilde{E}_{k}\|^{2}
            \end{equation}
            But at the same time:
            \begin{equation}\label{eq:error_m_k}
                \| \tilde{E}_{k+1}\| = \|M_{k+1} - I_{d} \| = \| M_{k}(2I_{d} - M_{k}) - I_{d}\| = \| -(M_{k} - I_{d})^{2}\| \leq \|\tilde{E}_{k}\|^{2}
            \end{equation}
            thus iterating Eq.~\eqref{eq:error_m_k}, we have that for all $k\in \mathbb{N}$, $\|\tilde{E}_{k}\| \leq \|\tilde{E}_{0}\|^{2^{k}}$.
            And then $\|\tilde{E}_{k}\| \leq \|\tilde{E}_{0}\|^{2^{k}}\|V^{-1}\|$, therefore using that any $V$ symmetric definite positive
            $\| V^{-1}\| = \|V\|^{-1}$ then for all $k\in \mathbb{N}$:
            \begin{equation}\label{eq:error_v}
                \|E_{k}\| \leq \left\| \frac{V}{c} - I_{d}\right\|^{2^{k}}\|V\|^{-1}
            \end{equation}
            But $\|\tilde{E}_{0}\| = \left\| \frac{1}{c}V - I_{d} \right\| = \max_{i\in[d]} \left| \frac{\lambda_{i}}{c} - 1\right|$
            where $\lambda_{1} \geq \lambda_{2} \geq \hdots \geq \lambda_{d}\geq 0$ are the (ordered) eigenvalues of $V$. However $c \geq  \text{Tr}(V)$ thus $0 \leq \lambda_{i}/c \leq 1$ for all $i\leq d$.
            Therefore $\|\tilde{E}_{0}\| \leq 1 - \frac{\lambda_{d}}{c}$. We also have that $\| V\| = \lambda_{1}$. Using Eq.~\eqref{eq:error_v}, we have for all $k$:
            \begin{align}\label{eq:last_bound_error}
                \| E_{k} \| \leq \left(1 - \frac{\lambda_{d}}{c}\right)^{2^{k}}\lambda_{1}^{-1} \leq \left(1 - \frac{\lambda}{c}\right)^{2^{k}}\lambda^{-1}
            \end{align}
            for any $0 \geq \lambda \leq \lambda_{d}$. Finally, Eq.~\eqref{eq:last_bound_error} implies that $\| E_{k}\| \leq \varepsilon$ as soon as:
            \begin{align}
                k\geq \frac{1}{\ln(2)}\ln\left( \frac{\ln(\lambda) + \ln(\varepsilon)}{\ln\left(1 - \frac{\lambda}{c} \right)}\right)
            \end{align}
            for any $0 \geq \lambda \leq \lambda_{d}$ and $\lambda \varepsilon \leq 1$.
        \end{proof}

        \subsection{Computing an Approximate Square Root}\label{app:proof_sqrt}

         The proof of Prop.~\ref{prop:iterations_square_root} is very similar to the proof of Prop.~\ref{prop:convergence_speed_approx_inv}
            thanks the analysis of the convergence speed in \cite{cheon2019numerical}. First, let us recall Prop.~\ref{prop:iterations_square_root}.

            \begin{proposition*}
                For any $z \in \mathbb{R}_{+}$, $c_{1},c_{2}>0$ with $c_{2} \geq z\geq c_{1}$
                and a precision $\varepsilon>0$, let $q_{k}$ be the result of $k$ iterations
                of Eq.~\eqref{eq:iterate_sqrt}, with $q_{0} = \frac{z}{c_{2}}$
                and $v_{0} = \frac{z}{c_{2}} - 1$.
                Then, $|q_{k}\sqrt{c_{2}} - \sqrt{z}| \leq \varepsilon$ for any
                $k\geq k_{0}(\varepsilon) := \frac{1}{\ln(2)}\ln\left(\frac{\ln\left(\varepsilon\right)-
                    \ln\left(\sqrt{c_{2}}\right)}{4\ln\left(1 - \frac{c_{1}}{4c_{2}}\right)}\right)$.
            \end{proposition*}

            \begin{proof}{of Prop.~\ref{prop:iterations_square_root}.} Because $0 \leq c_{1} < x < c_{2}$, we have that $\frac{x}{c_{2}} \in (0,1)$,
                hence thanks to Lemma $2$ of \cite{cheon2019numerical}, we have that after
                $k$ iterations:
                \begin{align}
                    \left|q_{k} - \sqrt{\frac{x}{c_{2}}}\right| \leq \left(1 - \frac{x}{4c_{2}}\right)^{2^{k+1}}
                \end{align}
                where $q_{k}$ is the $k$-th iterate from iterating Eq.~\eqref{eq:iterate_sqrt} with
                $q_{0} = \frac{x}{c_{2}}$ and $v_{0} = q_{0} - 1$.  Then because $x\geq c_{1}$,
                we have that $1 - \frac{x}{4c_{2}} \leq  1 - \frac{c_{1}}{4c_{2}}$. Stated otherwise,
                \begin{equation}
                    \left|q_{k} - \sqrt{\frac{x}{c_{2}}}\right|\leq \left(1 - \frac{c_{1}}{4c_{2}}\right)^{2^{k+1}}
                \end{equation}
                Therefore, for $k\geq \frac{1}{\ln(2)}\ln\left(\frac{\ln\left(\varepsilon\right)- \ln\left(\sqrt{c_{2}}\right)}{2\ln\left(1 - \frac{c_{1}}{4c_{2}}\right)}\right)$, the result follows since:
                \begin{equation}
                    \sqrt{c_{2}}\left|q_{k} - \sqrt{\frac{x}{c_{2}}}\right| \leq \varepsilon
                \end{equation}

            \end{proof}

        \subsection{Computing an Optimistic Ellipsoid Width.}\label{app:proof_distance_ols}

            The next step to build an optimistic algorithm is to compute a confidence ellipsoid around the
            estimate $\wt{\theta}_{t}$ such that the true parameter $\theta^{\star}$ belongs to this confidence
            ellipsoid with high probability. First, we need an estimate of the distance between $\theta^{\star}$ and $\wt{\theta}_{t}$ that is the object of
            Cor.~\ref{cor:distance_approx_ols}. The proof of Cor.~\ref{cor:distance_approx_ols}, is based on the fact that the approximated inverse is closed enough
            to the true inverse. Let's recall Cor.~\ref{cor:distance_approx_ols} first.

            \begin{corollary*}
                Setting $\varepsilon_t = \Big(Lt^{3/2}\sqrt{L^{2}t + \lambda}\Big)^{-1}$ in Prop.~\ref{prop:convergence_speed_approx_inv},
                then $\| \textit{Dec}_{\textbf{sk}}(\omega_{t}) - \theta_t\|_{V_{t}} \leq t^{-1/2}$, $\forall t$.
            \end{corollary*}

            \begin{proof}{of Cor.~\ref{cor:distance_approx_ols}.} Let's note $\bar{A}_{t}$, the result of iterating Eq.~\eqref{eq:iterate_inverse}, $k_{1}(\varepsilon_{t})$ times
                with $V = V_{t}$ and $c = \lambda d + L^{2}t$.
                Thanks to the definition of $\text{Dec}_{\textbf{sk}}(w_{t})$ and $\theta_{t} = V_{t}^{-1}b_{t}$, we have:
				\begin{align}
					\|\text{Dec}_{\textbf{sk}}(w_{t}) - \theta_{t}\|_{V_{t}} &= \left\|V_{t}^{1/2}\left(V_{t}^{-1} - \text{Dec}_{\textbf{sk}}(\bar{A}_{t})\right) \sum_{l=1}^{t-1} r_{l}s_{l,a_{l}}\right\|_{2}\\
					&= \left\|\left(V_{t}^{-1} - \text{Dec}_{\textbf{sk}}(\bar{A}_{t})\right) V_{t}^{1/2}\sum_{l=1}^{t - 1} r_{l}s_{l,a_{l}}\right\|_{2}\\
					&\leq \| \text{Dec}_{\textbf{sk}}(\bar{A}_{t}) - V_{t}^{-1}\| \left\| V_{t}^{1/2}\sum_{l=1}^{t-1} r_{l}s_{l,a_{l}}\right\|_{2}
				\end{align}
                But $\text{Tr}(V_{t}) \leq \lambda d + L^{2}t$ and $\lambda_{\min}(V_{t}) \geq \lambda$.
				Therefore thanks to Prop.~\ref{prop:convergence_speed_approx_inv} $\bar{A}_{t}$ is such that:
				\begin{align}
					\| \text{Dec}_{\textbf{sk}}(\bar{A}_{t}) - V_{t}^{-1} \| \leq \varepsilon_{t}
				\end{align}
                We also have that:
				\begin{align}
					\left\| V_{t}^{1/2}\sum_{l=1}^{t-1} r_{l}s_{l,a_{l}}\right\|_{2}&\leq \| \sqrt{V_{t}}\|\left\|\sum_{l=1}^{t-1} r_{l}s_{l,a_{l}}\right\|_{2}\\
					&\leq Lt \sqrt{\|V_{t}\|} \leq Lt\sqrt{\lambda + L^{2}t}
				\end{align}
                because $r_{l} \in [-1,1]$ for all $l\leq t$ and $\lambda_{\max}(V_{t}) \leq \lambda + L^{2}t$.
                Finally, we have that:
                \begin{align}
                    \|\theta_{t} - \tilde{\theta}_{t}\|_{V_{t}} \leq \varepsilon_{t} Lt\sqrt{\lambda + L^{2}t} \leq t^{-1/2}
                \end{align}
            \end{proof}

        \subsection{Approximate Confidence Ellipsoid}\label{app:confidence_theta}

            Finally thanks to Cor.~\ref{cor:distance_approx_ols}, we can now prove that with high probability $\theta^{\star}$
            belongs to the inflated confidence intervals $\tilde{\mathcal{C}}_{t}$ for all time $t$.
            That is the object of Prop.~\ref{prop:confidence_theta}.

            \begin{proposition}\label{prop:confidence_theta}
                For any $\delta>0$, we have that with probability at least $1 - \delta$:
                \begin{align}
                    \theta^{\star} \in \bigcap_{t=1}^{+\infty} \mathcal{C}_{t}(\delta) := \left\{ \theta \mid \left\| \theta - \text{Dec}_{\textbf{sk}}(w_{t}) \right\|_{V_{t}} \leq \wt{\beta}(t)\right\}
                \end{align}
                with $\wt{\beta}(t) = t^{-1/2} + \sqrt{\lambda}S + \sigma\sqrt{d(\ln(1 + L^{2}t/(\lambda d)) +  \ln(\pi^{2}t^{2}/(6\delta))}$
            \end{proposition}

            \begin{proof}{of Prop.~\ref{prop:confidence_theta}.} Using Cor.~\ref{cor:distance_approx_ols} and Thm.~$2$ in \cite{abbasi2011improved}, we have that for any time $t$ that with probability at least $1 - \delta$:
                    \begin{align}
                        \| \theta^{\star} - \text{Dec}_{\textbf{sk}}(w_{t})\|_{V_{t}} &\leq \| \theta_{t} - \text{Dec}_{\textbf{sk}}(w_{t})\|_{V_{t}}  + \| \theta^{\star} - \theta_{t}\|_{V_{t}}\\
                        &\leq t^{-1/2} + \sqrt{\lambda} S + \sigma\sqrt{d(\ln(1 + L^{2}t/(\lambda d)) +  \ln(1/\delta))}
                    \end{align}
                    where $w_{t}$ computed as in Alg.~\ref{alg:low_switching_algo} and $\theta_{t}$ is the ridge regression estimate computed
                    at every time step in \oful. Taking a union bound with high-probability event means that with probability at least $1 - \frac{6\delta}{\pi^{2}}$, we have:
                    \begin{align}
                        \| \theta^{\star} - \theta_{t}\|_{V_{t}} &\leq \| \theta_{t} - \text{Dec}_{\textbf{sk}}(w_{t})\|_{V_{t}}  + \| \theta^{\star} - \theta_{t}\|_{V_{t}}\\
                        &\leq t^{-1/2} + \sqrt{\lambda} S + \sigma\sqrt{d(\ln(1 + L^{2}t/(\lambda d)) +  \ln(\pi^{2}t^{2}/(6\delta)))}
                    \end{align}
            \end{proof}

        \subsection{Homomorphic Friendly Approximate Argmax}\label{app:computing_argmax}

            As mentioned in Sec.~\ref{sec:algo}, an homomorphic algorithm can not directly
            compute the argmax of a given list of values.  In this work, we introduce the algorithm
            Alg.~\ref{alg:newton_comp} to compute the comparison vector $b_{t} \approxeq \left(\mathds{1}_{\{a = \arg\max_{i\in [K]} \rho_{i}(t)\}}\right)$
            with $(\rho_{a}(t))_{a\in[K]}$ the UCBs defined in Sec.~\ref{sec:algo}. This algorithm
            is divided in two parts. First, it computes an approximate maximum, $M$ of $(\rho_{a}(t))_{a\in[K]}$
            thanks to Alg.~\ref{alg:newton_max} and then compares each values $(\rho_{a}(t))_{a\in[K]}$ to this
            approximate maximum $M$ thanks to the algorithm $\textbf{NewComp}$ of \cite{cheon2019efficient} (recalled as Alg.~\ref{alg:new_comp}).

            \begin{algorithm}[h]
                \caption{NewComp}
                \label{alg:new_comp}
                \begin{algorithmic}
                    \STATE {\bfseries Input:} Entry numbers: $a,b\in[0,1]$, $n$ and depth $d$
                    \STATE Set $x =  a-b$
                    \FOR{$k=1,\dots, d$}
                        \STATE Compute $x = f_{n}(x) = \sum_{i=0}^{n} \frac{1}{4^{i}}\binom{2i}{i}x(1 - x^{2})^{i}$
                    \ENDFOR
                    \STATE {\bfseries Return:} $(x+1)/2$
                \end{algorithmic}
            \end{algorithm}

            \begin{algorithm}[h]
                \caption{NewMax}
                \label{alg:new_max}
                \begin{algorithmic}
                    \STATE {\bfseries Input:} Entry numbers: $a,b\in[0,1]$, $n$ and depth $d$
                    \STATE Set $x = a-b$, $y = \frac{a+b}{2}$
                    \FOR{$k=1,\dots, d$}
                        \STATE Compute $x = f_{n}(x) = \sum_{i=0}^{n} \frac{1}{4^{i}}\binom{2i}{i}x(1 - x^{2})^{i}$
                    \ENDFOR
                    \STATE {\bfseries Return:} $y + \frac{a+b}{2}\cdot x$
                \end{algorithmic}
            \end{algorithm}


            \begin{algorithm}[h]
                \caption{amax}
                \label{alg:newton_max}
                \begin{algorithmic}
                    \STATE {\bfseries Input:} Entry numbers: $(a_{i})_{i\leq K}$, $n$ and depth $d$
                    \STATE Set $m =  a_{1}$
                    \FOR{$i=2,\dots, K$}
                        \STATE Compute $m = \max\{m, a_{i}\}$ thanks to \textbf{NewMax}
                        in \cite{cheon2019efficient} with parameter $a = m$, $b = a_{i}$, $n$ and $d$
                    \ENDFOR
                \end{algorithmic}
            \end{algorithm}
            \begin{algorithm}[h]
                \caption{acomp}
                \label{alg:newton_comp}
                \begin{algorithmic}
                    \STATE {\bfseries Input:} Entry numbers: $(a_{i})_{i\leq K}$, precision $\varepsilon$
                    \STATE Set depth $d = 1 + \left\lfloor 3.2 + \frac{\ln(1/\varepsilon)}{\ln(3/2)} + \frac{\ln\left(\frac{\ln\left(1/\varepsilon\right)}{\ln(2)} - 2\right)}{\ln(2)} \right\rfloor$ and depthmax $d' = \frac{1}{\ln(3/2)}\ln\left(\frac{\alpha\ln\left( \frac{1}{\varepsilon}\right)}{\ln(2)} - 2 \right)$
                    with $\alpha =  \frac{3}{2} + \frac{5.2\ln(3/2)}{\ln(4)} + \frac{\ln(3/2)}{2\ln(2)}$
                    \STATE Compute $M = \text{amax}((a_{i})_{i\leq K}, n, d)$
                    \FOR{$i=2,\dots, K$}
                        \STATE Set $b_{i} = \text{\textbf{NewComp}}(a_{i}, M, n ,d')$
                    \ENDFOR
                \end{algorithmic}
            \end{algorithm}

            \paragraph{Rescaling the UCB index:}

                In order to use the HE-friendly algorithms of \cite{cheon2019efficient}, we need to rescale the UCB-index to lies in $[0,1]$. Determining the
                range of those indexes is the purpose of the following proposition.

                \begin{proposition}\label{prop:bound_ucb}
                    For every time $t\geq 1$, assuming $r_l \in [-1,1]$ for any $l\leq t$ and $L\geq 1$ then for any $\delta>0$ we have that
                    with probability at least $1 - \delta$:
                    \begin{align}
                        -1 \leq \rho_{a}(t) \leq 1 + 2\wt{\beta}(t)\left[2t^{-1} + L\sqrt{\frac{1}{\lambda} + \frac{1}{\sqrt{\lambda}}} +  \sqrt{\frac{L}{t^{3/2}\sqrt{\lambda + L^{2}t}}}\right]
                    \end{align}
                    where $\rho_{a}(t) = \langle \wt{\theta}_{t}, x_{t,a}\rangle + \wt{\beta}(t)\left[q_{k_{0}(t^{-1})} + t^{-1}\right]$ the UCB index of arm $a$ at time $t$.
                \end{proposition}

                \begin{proof}{of Prop.~\ref{prop:bound_ucb}.}
                    For $\delta >0$, we denote $E = \bigcap_{l = 1}^{+\infty} \left\{ \theta^{\star} \in \tilde{\mathcal{C}}_{l}(\delta)\right\}$
                    so that, using Prop.~\ref{prop:confidence_theta}, $\mathbb{P}(E) \geq 1- \delta$.
                    Under the event $E$, we have for any arm $a$:
                    \begin{align}
                        -1 \leq \langle x_{t,a}, \theta^{\star} \rangle \leq \rho_{a}(t) \leq \langle x_{t,a}, \theta^{\star} \rangle + 2\wt{\beta}(t)\left[q_{k_{0}(1/t)} +t^{-1}\right]
                    \end{align}
                    On the other hand thanks to Prop.~\ref{prop:iterations_square_root}, we have that $q_{k_{0}(t^{-1})} \leq \sqrt{\| x\|_{\text{Dec}_{\textbf{sk}}(A_{t})}^{2} + \frac{L}{t^{3/2}\sqrt{\lambda + L^{2}t}}} + t^{-1}$.
                    and also $\| x\|_{A_{t}}^{2} \leq L^{2}\left(\frac{1}{\lambda} + \frac{1}{\sqrt{\lambda}}\right)$.

                    Indeed because $\text{Dec}_{\textbf{sk}}(A_{t})$ is a polynomial function of $V_{t}$, we have that $\text{Dec}_{\textbf{sk}}(A_{t})$ is
                    symmetric and $\text{Dec}_{\textbf{sk}}(A_{t})V_{t} = V_{t}\text{Dec}_{\textbf{sk}}(A_{t})$, hence $\text{Dec}_{\textbf{sk}}(A_{t})$ and $V_{t}^{-1}$ are diagonalizable
                    in the same basis therefore $\|\text{Dec}_{\textbf{sk}}(A_{t}) - V_{t}^{-1}\| = \max_{i\leq d} | \lambda_{i}(\text{Dec}_{\textbf{sk}}(A_{t})) - \lambda_{i}(V_{t}^{-1})|$
                    with $\lambda_{i}(M)$ the $i$-th biggest eigenvalue of $M$. Hence:
                    \begin{align}
                    	\lambda_{1}(\text{Dec}_{\textbf{sk}}(A_{t})) \leq \frac{1}{\lambda} + \frac{1}{Lt^{3/2}\sqrt{\lambda + L^{2}t}}
                    \end{align}
                    and:
                    \begin{align}
                    	\lambda_{d}(\text{Dec}_{\textbf{sk}}(A_{t})) \geq \frac{1}{\lambda + L^{2}t} - \frac{1}{Lt^{3/2}\sqrt{\lambda + L^{2}t}} > 0
                    \end{align}
                    for $t\geq 2$. Therefore, we have that for any arm $a$:
                    \begin{align}
                        \rho_{a}(t) \leq \langle \theta^{\star}, x_{t,a}\rangle + 2\wt{\beta}(t)\left[2t^{-1} + L\sqrt{\frac{1}{\lambda} + \frac{1}{\sqrt{\lambda}}} +  \sqrt{\frac{L}{t^{3/2}\sqrt{\lambda + L^{2}t}}}\right]
                    \end{align}
                \end{proof}

            \paragraph{Computing the Comparaison Vector:}
                The algorithm Alg.~\ref{alg:newton_comp} operates on values in $[0,1]$ therefore using Prop.~\ref{prop:bound_ucb}, we can
                compute rescaled UCB index, noted $\tilde{\rho}_{a}(t) \in [0,1]$. We are then almost ready to prove Cor.~\ref{cor:precision_argmax},
                we just need two lemmas which relates the precision of Alg.~\ref{alg:newton_comp} and Alg.~\ref{alg:newton_max} to the precision
                of $\textbf{NewComp}$ and $\textbf{NewMax}$ of \cite{cheon2019efficient}.

                The first lemma (Lem.~\ref{lemma:max_precision}) gives a lower bound on the depth needed for Alg.~\ref{alg:newton_max}
                to achieve a given precision.

                \begin{lemma}\label{lemma:max_precision}
					For any sequences $(a_{i})_{i\leq K}\in [0,1]^{K}$, for any precision $0<\varepsilon<K/4$, $n\in \mathbb{N}^{\star}$ and
					\begin{align}
						d(\varepsilon, n) \geq \frac{\ln\left(\frac{\ln\left(\frac{K}{\varepsilon}\right)}{\ln(2)} - 2\right)}{\ln(c_{n})}
					\end{align}
					with $c_{n} = \frac{2n+1}{4^{n}}\binom{2n}{n}$. Noting $M$ the result of Alg.~\ref{alg:newton_max} with parameter $(a_{i})_{i}$, $n$ and $d(\varepsilon, n)$, we have that:
					\begin{align}
						\left| M - \max_{i} a_{i}\right| \leq \varepsilon
					\end{align}
                \end{lemma}

                \begin{proof}{of Lemma~\ref{lemma:max_precision}.}
					Thanks to Corollary $4$ in \cite{cheon2019efficient}, we have that for any $n$ and depth $d\geq \frac{\ln\left(\frac{\ln(1/\varepsilon)}{\ln(2)} - 2\right)}{\ln(c_{n}) }$ (with $c_{n} = \frac{2n+1}{4^{n}}\binom{2n}{n}$) and number $a,b$:
					\begin{align}
						\left| \text{\textbf{NewMax}}(a, b, n, d) - \max\{a, b\} \right| \leq  \varepsilon
					\end{align}

                    Let's note $m_{k}$ the iterate $m$ of Alg.~\ref{alg:newton_max} at step $k\in[K]$ in the for loop. We show that by induction
                    $\left|m_{k} - \max_{i\in[k]} a_{i} \right| \leq k\varepsilon$.
                    \begin{itemize}
                        \item By definition $m_{1} = a_{1}$ and $\left|m_{1} - \max_{i\leq 1} a_{i}\right| = 0$
                        \item Using that $\left|\max\{a, c\} - \max\{b,c\}\right|\leq \left| a - b\right|$ for any $a,b,c\in\mathbb{R}$, we have:
                                {\small\begin{align*}
                                    \left| m_{k+1} - \max_{i\leq k+1} a_{i} \right| &= \Big| \text{\textbf{NewMax}}(m_{k}, a_{k+1}, n, d) - \max\{ m_{k}, a_{k+1}\} \\
                                    &\hspace{3cm}+ \max\{ m_{k}, a_{k+1}\} - \max\{\max_{i\leq k} a_{i}, a_{k+1}\}\Big|\\
                                    &\leq | \text{\textbf{NewMax}}(m_{k}, a_{k+1}, n, d) - \max\{ m_{k}, a_{k+1}\}| \\
                                    &\hspace{3cm}+ |\max\{ m_{k}, a_{k+1}\} - \max\{\max_{i\leq k} a_{i}, a_{k+1}\}|\\
                                    &\leq \varepsilon + |m_{k} - \max_{i\leq k} a_{i}| \leq (k+1) \varepsilon
                                \end{align*}}
                    \end{itemize}
					Finally, because $M = m_{K}$, we just need to choose $d\geq \frac{\ln\left(\frac{\ln(K/\varepsilon)}{\ln(2)} - 2\right)}{\ln(c_{n})}$ to get the result.
                \end{proof}

                The next lemma (Lem.~\ref{lemma:comp_lemma}) has the same purpose of Lem.~\ref{lemma:max_precision} but this time for Alg.~\ref{alg:newton_comp}.
                The proof is based on properties of the polynomial function used by the algorithm $\textbf{NewComp}$ in order to predict the
                result of the comparison when the margin condition of $\textbf{NewComp}$ (that is to say the result of the comparison of
                $a,b\in [0,1]$ is valid if and only if $|a -b|\geq \varepsilon$ for some $\varepsilon>0$) is not satisfied.

                \begin{lemma}\label{lemma:comp_lemma}
                    For $\varepsilon\in (0, 1/4)$ and sequence $(a_{i})_{i\leq K}\in [0,1]^{K}$, let's denote $(b_{i})_{i\leq K}$ te result of Alg.~\ref{alg:newton_comp} ruuned with
                    parameter $(a_{i})_{i\leq K}$, $n = 1$, $d' = d_{2}(\varepsilon)$ and $d = d_{3}(\varepsilon)$ with:
                    \begin{align}
                        &d_{2}(\varepsilon) =  \left\lfloor3.2 + \frac{\ln(1/\varepsilon)}{\ln(c_{n})} + \frac{\ln\left( \ln\left(\frac{1}{\varepsilon}\right)/\ln(2)-2\right)}{\ln(n+1)}\right\rfloor + 1\\
                        &d_{3}(\varepsilon) \geq \frac{1}{\ln(c_{n})}\ln\left(\frac{\alpha\ln\left( \frac{1}{\varepsilon}\right)}{\ln(2)} - 2 \right)
                    \end{align}
                    where $\alpha = \frac{3}{2} + \frac{5.2\ln(c_{n})}{\ln(4)} + \frac{\ln(c_{n})}{2\ln(n+1)}$. Then selecting any $i\leq K$ such that $b_{i} \geq \varepsilon$ (and there is at least one such index $i$), we have that $a_{i} \geq \max_{k} a_{k} - 2\varepsilon$
                \end{lemma}

                \begin{proof}{of Lemma~\ref{lemma:comp_lemma}.} Thanks to Corollary $1$ in \cite{cheon2019numerical}, we have that for each $i\leq K$, $\left| b_{i} - \text{Comp}(a_{i}, M)\right| \leq \varepsilon$ as soon as
                    $\left|a_{i} - M \right| > \varepsilon$ and $d' =  \left\lfloor3.2 + \frac{\ln(1/\varepsilon)}{\ln(c_{n})} + \frac{\ln\left( \ln\left(\frac{1}{\varepsilon}\right)/\ln(2)-2\right)}{\ln(n+1)}\right\rfloor + 1	$.
                    For $i\in [K]$, we have that:
                    \begin{itemize}
                        \item If $\max_{k\leq K} a_{k} \geq a_{i}\geq M + \varepsilon$ then $\text{Comp}(a_{i}, M) = 1$, $|b_{i} - 1|\leq \varepsilon$ and
                        $ a_{i} \geq \max_{k\leq K}a_{k} - |\max_{k\leq K}a_{k} - M|- \varepsilon$
                        \item If $a_{i}\leq M - \varepsilon$ then $\text{Comp}(a_{i}, M) = 0$, thus $|b_{i}|\leq \varepsilon$ and
                        $ a_{i} \leq \max_{k\leq K}a_{k} + |\max_{k\leq K}a_{k} - M|- \varepsilon$
                    \end{itemize}
                    Therefore for any $a_{i}$ such that $|a_{i} - M| > \varepsilon$ then the resulting $b_{i}$ is either bounded by $1-\varepsilon$ or $\varepsilon$.

                    The second option is if $|a_{i} - M| \leq \varepsilon$ then the \textbf{NewComp} algorithm provides no guarantee to the result of the
                    algorithm. However the algorithm applies a function $f_{n}$\footnote{For all $x\in [-1,1], f_{n}(x) = \sum_{i=0}^{n} \frac{1}{4^{i}}\binom{2i}{i}x(1 - x^{2})^{i}$.} multiple times to its input.
                    For every $x\in [-1,1]$:
                    \begin{align}
                            |f_{n}(x)| \leq c_{n}|x| \text{ and } f_{n}([-1,1]) \subset [-1,1]
                    \end{align}
                    with $c_{n} = \frac{2n+1}{4^{n}}\binom{2n}{n}$.
                    Hence:
                    \begin{align}
                        \forall x\in [-1,1] \qquad |f_{n}^{(d')}(x)| \leq c_{n}|f_{n}^{(d'-1)}(x)| \leq c_{n}^{d"}|x|
                    \end{align}
                    But if $|a_{i} - M|\leq \varepsilon$, $f_{n}^{(d')}(a_{i} - M) \leq c_{n}^{d'}|a_{i} - M| \leq c_{n}^{d'}\varepsilon$ thus
                    $\left|b_{i} - \frac{1}{2}\right| \leq \frac{c_{n}^{d'}\varepsilon}{2}$.

                    Finally for each $i$, we only three options for $b_{i}$:
                    \begin{itemize}
                        \item If $|a_{i} - M| \leq \varepsilon$ then $\left| b_{i} - \frac{1}{2}\right| \leq \frac{c_{n}^{d'}\varepsilon}{2}$ and $a_{i} \geq \max_{k} a_{k} - (\varepsilon + |\max_{k} a_{k} - M|)\geq  \max_{k} a_{k} - 2\varepsilon$
                        \item If $|a_{i} - M| \geq \varepsilon$ and $a_{i} \leq  M - \varepsilon$ then $|b_{i}| \leq \varepsilon$ and $a_{i} \leq \max_{k} a_{k} + |M - \max_{k} a_{k}| - \varepsilon \leq \max_{k} a_{k}$
                        \item If $|a_{i} - M| \geq \varepsilon$ and $a_{i} \geq  M + \varepsilon$ then $|b_{i}-1| \leq \varepsilon$ and $a_{i} \geq \max_{k} a_{k} - (|M - \max_{k} a_{k}| + \varepsilon)\geq \max_{k} a_{k} - 2\varepsilon$
                    \end{itemize}
                    To finish the proof, we just need to ensure that there exists at least one $i$ such that $b_{i} \geq \varepsilon$. Noting $i^{\star} = \arg\max_{k} a_{k}$, if the $\text{amax}$ algorithm is used with depth $d$ such that:
                    \begin{align}
                        d\geq \frac{1}{\ln(c_{n})}\ln\left(\frac{\alpha\ln\left( \frac{1}{\varepsilon}\right)}{\ln(2)} - 2 \right)
                    \end{align}
                    where $\alpha = \frac{3}{2} + \frac{5.2\ln(c_{n})}{\ln(4)} + \frac{\ln(c_{n})}{2\ln(n+1)}$, we have $|a_{i^{\star}} - M| \leq \varepsilon^{\alpha} \leq \varepsilon$ and $b_{i^{\star}} \geq \frac{1}{2} - \frac{c_{n}^{d'}\varepsilon^{\alpha}}{2} > \varepsilon$. Hence there always exists an index $i$ such that $b_{i} \geq \varepsilon$.
                \end{proof}

                Finally, thanks to Lem.~\ref{lemma:comp_lemma}, we can finally  prove Cor.~\ref{cor:precision_argmax}. The proof of this corollary
                simply amounts to choose the right precision for $\textbf{NewComp}$ algorithm at every step of Alg.~\ref{alg:newton_comp}. First let's recall
                Cor.~\ref{cor:precision_argmax}.

                \begin{corollary*}
                    For any time $t$, selecting any arm $a$ such that $(b_{t})_{a} \geq \frac{1}{4t}$ then:
                    \begin{equation}
                        \begin{aligned}
                        &\rho_{a}(t) \geq \max_{k\leq K} \rho_{k}(t) - \frac{1}{t}\left(1 + \wt{\beta}^{\star}(t)\right)
                        \end{aligned}
                    \end{equation}
                    where $\wt{\beta}^{\star}(t) = \wt{\beta}(t)\left[2t^{-1} +  \sqrt{\frac{L}{t^{3/2}\sqrt{\lambda + L^{2}t}}} + L\sqrt{\frac{1}{\lambda} + \frac{1}{\sqrt{\lambda}}}\right]$
                \end{corollary*}

                \begin{proof}{of Cor.~\ref{cor:precision_argmax}.} Using Lem.~\ref{lemma:comp_lemma} with $\varepsilon = \frac{1}{4t}$ yields the following result:
                    \begin{align}
                        \tilde{\rho}_{i}(t) \geq \max_{k\leq K} \tilde{\rho}_{k}(t) - \frac{1}{2t}
                    \end{align}
                    But for any $i\leq K$, $\tilde{\rho}_{i}(t) = (\rho_{i}(t) +1)/\left(2 + 2\wt{\beta}(t)\left[2t^{-1} +  \sqrt{\frac{L}{t^{3/2}\sqrt{\lambda + L^{2}t_{j}}}} + L\sqrt{\frac{1}{\lambda} + \frac{1}{\sqrt{\lambda}}}\right]\right)$. Hence the result.
                \end{proof}

\section{SLOW SWITCHING CONDITION AND REGRET OF \algo}

    In this appendix, we present the analysis of the regret of \secofulls. The proof is decomposed in two steps.
    The first one is the analysis of the number of batches for any time $T$. That is the object of the Sec.~\ref{app:proof_nb_batches}.
    The second part of the proof amounts to bounding the regret as a function of the number of batches (Sec.~\ref{app:proof_regret_efficient_linucb}).

    \subsection{Number of batches of \secofulls (Proof of Prop.~\ref{prop:nb_batches})}\label{app:proof_nb_batches}

        We first prove Prop.~\ref{prop:nb_batches} which states that the total number of batches for \secofulls is
        logarithmic in $T$ contrary to \secoful where the parameter are updated a linear number of times. The proof of this proposition
        is itself divided in multiple steps. First, we show how using $\textbf{NewComp}$ to compare the parameter $C$ and $\text{Tr}\left( \text{Dec}_{\textbf{sk}}(\bar{A}_{j})\sum_{l=t_j+1}^{t-1} s_{l,a_{l}}s_{l,a_{l}}^{\intercal} \right)$ (for any batch $j$)
        relate to the comparison of $C$ and $\text{Tr}\left( \bar{V}_{j}^{-1}\sum_{l=t_j+1}^{t-1} s_{l, a_{l}}s_{l, a_{l}}^{\intercal} \right)$. Then, we show how Condition~\ref{eq:trace_condition}
        relates to the det-based condition used in \rsoful which allows us to finish the proof of Prop.~\ref{prop:nb_batches} following the
        same reasoning as in \cite{abbasi2011improved}.

        \subsubsection{Homomorphically Friendly Comparison for Condition~\ref{eq:trace_condition}}\label{app:trace_condition}

            We first prove the following proposition, bounding the error made by our algorithm when using
            $ \text{Tr}\left(\text{Dec}_{\textbf{sk}}(\bar{A}_{j})\sum_{l=t_j +1}^{t-1} s_{l, a_{l}}s_{l, a_{l}}^{\intercal} \right)$  instead of
            $ \text{Tr}\left( \bar{V}_{j}^{-1}\sum_{l=t_j+1}^{t-1} s_{l, a_{l}}s_{l, a_{l}}^{\intercal} \right)$.

            \begin{proposition}\label{prop:precision_comparaison_trace}
                For an batch $j$, time $t\geq t_{j} + 1$, $\varepsilon<1/2$ and $\varepsilon'>0$, let's note $\delta_{t}$ the result of $\textbf{NewComp}$ applied with parameters
                $a = \frac{\text{Tr}\left( \text{Dec}_{\textbf{sk}}(\bar{A}_{j}) \sum_{l=t_{j}+1}^{t-1} s_{l, a_{l}}s_{l, a_{l}}^{\intercal}\right)}{L^{2}\left( \frac{1}{\lambda} + \frac{1}{\sqrt{\lambda}}\right)(t-1-t_{j})}$, $b= \frac{C}{{L^{2}\left( \frac{1}{\lambda} + \frac{1}{\sqrt{\lambda}}\right)(t-1-t_{j})}}$ \footnote{with the convention that $0/0 = 0$ and $C/0 = 1$},
                $n=1$ and $d_{5}(\varepsilon)$ such that:
                \begin{align}
                    d_{5}(\varepsilon) \geq 3.2 + \frac{\ln(1/\varepsilon')}{\ln(c_{n})} + \frac{\ln\left(\ln\left(\frac{1}{\varepsilon}\right)/\ln(2) - 2\right)}{\ln(n+1)}
                \end{align}
                then:
                \begin{itemize}
                    \item if $\delta_{t} > \varepsilon$:
                        \begin{align}
                            C - \varepsilon' L^{2}\left( \frac{1}{\lambda} + \frac{1}{\sqrt{\lambda}}\right)(t-1-t_{j}) \leq \text{Tr}\left(\text{Dec}_{\textbf{sk}}(\bar{A}_{j})\sum_{l=t_j +1}^{t-1} s_{l, a_{l}}s_{l, a_{l}}^{\intercal} \right)
                        \end{align}
                    \item else if $\delta_{t} \leq \varepsilon$:
                        \begin{align}
                            \text{Tr}\left(\text{Dec}_{\textbf{sk}}(\bar{A}_{j})\sum_{l=t_j +1}^{t-1} s_{l, a_{l}}s_{l, a_{l}}^{\intercal} \right) \leq C + \varepsilon'L^{2}\left( \frac{1}{\lambda} + \frac{1}{\sqrt{\lambda}}\right)(t-1-t_{j})
                        \end{align}
                    \end{itemize}
            \end{proposition}
            \begin{proof}{of Prop.~\ref{prop:precision_comparaison_trace}.} We consider the two cases, depending if $\delta_{t}$ is bigger than $\varepsilon$
                or not.

                \textbf{If $\delta_{t}> \varepsilon$}: we proceed by separation of cases.
                    \begin{itemize}
                        \item If $\left|\text{Tr}\left(\text{Dec}_{\textbf{sk}}(\bar{A}_{j})\sum_{l=t_j +1}^{t-1} s_{l, a_{l}}s_{l, a_{l}}^{\intercal} \right) - C\right| > \varepsilon'L^{2}\left( \frac{1}{\lambda} + \frac{1}{\sqrt{\lambda}}\right)(t-1-t_{j})$:

                            $$\left| \delta_{t} - \text{Comp}\left(\text{Tr}\left(\text{Dec}_{\textbf{sk}}(\bar{A}_{j})\sum_{l=t_j +1}^{t-1} s_{l, a_{l}}s_{l, a_{l}}^{\intercal} \right), C\right)\right| \leq \varepsilon$$

                            thanks to Cor.~$1$ in \cite{cheon2019efficient} for the precision of $\textbf{NewComp}$. We also used the fact that for any $x,y \in \mathbb{R}$ and $z\in\mathbb{R}_{+}^{\star}$, $\text{Comp}(x/z,y/z) = \text{Comp}(x,y)$. Using the equation above:

                                $$\text{Comp}\left(\text{Tr}\left(\text{Dec}_{\textbf{sk}}(\bar{A}_{j})\sum_{l=t_j +1}^{t-1} s_{l, a_{l}}s_{l, a_{l}}^{\intercal} \right), C\right) \geq \delta_{t} - \varepsilon>0$$

                            because we assumed here that $\delta_{t} > \varepsilon$.
                            This readily implies that $\text{Comp}\left(\text{Tr}\left(\text{Dec}_{\textbf{sk}}(\bar{A}_{j})\sum_{l=t_j +1}^{t-1} s_{l, a_{l}}s_{l, a_{l}}^{\intercal} \right), C\right) = 1$ because $\text{Comp}(a,b) \in \{0, 1\}$ for any $a,b\in [0,1]$.
                            But, because we are in the case that:
                            $$\left|\text{Tr}\left(\text{Dec}_{\textbf{sk}}(\bar{A}_{j})\sum_{l=t_j +1}^{t-1} s_{l, a_{l}}s_{l, a_{l}}^{\intercal} \right) - C\right| > \varepsilon'L^{2}\left( \frac{1}{\lambda} + \frac{1}{\sqrt{\lambda}}\right)(t-1-t_{j})$$
                            we have that either $\text{Tr}\left(\text{Dec}_{\textbf{sk}}(\bar{A}_{j})\sum_{l=t_j +1}^{t-1} s_{l, a_{l}}s_{l, a_{l}}^{\intercal} \right) > C + \varepsilon'L^{2}\left( \frac{1}{\lambda} + \frac{1}{\sqrt{\lambda}}\right)(t-1-t_{j})$ or $\text{Tr}\left(\text{Dec}_{\textbf{sk}}(\bar{A}_{j})\sum_{l=t_j +1}^{t-1} s_{l, a_{l}}s_{l, a_{l}}^{\intercal} \right) < C -\varepsilon'L^{2}\left( \frac{1}{\lambda} + \frac{1}{\sqrt{\lambda}}\right)(t-1-t_{j})$.
                            Hence, because $\text{Comp}\left(\text{Tr}\left(\text{Dec}_{\textbf{sk}}(\bar{A}_{j})\sum_{l=t_j +1}^{t-1} s_{l, a_{l}}s_{l, a_{l}}^{\intercal} \right), C\right) = 1$, we have that $\text{Tr}\left(\text{Dec}_{\textbf{sk}}(\bar{A}_{j})\sum_{l=t_j +1}^{t-1} s_{l, a_{l}}s_{l, a_{l}}^{\intercal} \right)> C$ that is to say:
                            {\small
                            \begin{align*}
                                \text{Tr}\left(\text{Dec}_{\textbf{sk}}(\bar{A}_{j})\sum_{l=t_j +1}^{t-1} s_{l, a_{l}}s_{l, a_{l}}^{\intercal} \right) &\geq C + \varepsilon'L^{2}\left( \frac{1}{\lambda} + \frac{1}{\sqrt{\lambda}}\right)(t-1-t_{j}) \\
                                &\geq C - \varepsilon' L^{2}\left( \frac{1}{\lambda} + \frac{1}{\sqrt{\lambda}}\right)(t-1-t_{j})
                            \end{align*}}

                        \item If $\left|\text{Tr}\left(\text{Dec}_{\textbf{sk}}(\bar{A}_{j})\sum_{l=t_{j}+1}^{t-1} s_{l,a_{l}}s_{l,a_{l}}^{\intercal} \right) - C\right| \leq \varepsilon'L^{2}\left( \frac{1}{\lambda} + \frac{1}{\sqrt{\lambda}}\right)(t-1-t_{j})$:
                            We can not use Cor.~\ref{cor:precision_argmax} from \cite{cheon2019efficient}. However, in this case we directly have by definition of the absolute value that:
                            {\small\begin{equation}
                                - \varepsilon'L^{2}\left( \frac{1}{\lambda} + \frac{1}{\sqrt{\lambda}}\right)(t-1-t_{j}) \leq \text{Tr}\left(\text{Dec}_{\textbf{sk}}(\bar{A}_{j})\sum_{l=t_{j}+1}^{t-1} s_{l,a_{l}}s_{l,a_{l}}^{\intercal} \right) - C 
                            \end{equation}}
                    \end{itemize}

                \textbf{If $\delta_{t} \leq \varepsilon$:} Again, we distinguish the two different cases possible.
                    \begin{itemize}
                        \item If $\left|\text{Tr}\left(\text{Dec}_{\textbf{sk}}(\bar{A}_{j})\sum_{l=t_{j}+1}^{t-1} s_{l,a_{l}}s_{l,a_{l}}^{\intercal} \right) - C\right| > \varepsilon'L^{2}\left( \frac{1}{\lambda} + \frac{1}{\sqrt{\lambda}}\right)(t-1-t_{j})$:
                            Using Cor.~$1$ from \cite{cheon2019efficient}, we have once again that:
                                $$\left| \delta_{t} - \text{Comp}\left(\text{Tr}\left(\text{Dec}_{\textbf{sk}}(\bar{A}_{j})\sum_{l=t_{j}+1}^{t-1} s_{l,a_{l}}s_{l,a_{l}}^{\intercal}\right), C\right)\right| \leq \varepsilon$$
                            Therefore $\text{Comp}\left(\text{Dec}_{\textbf{sk}}(\bar{A}_{j})\sum_{l=t_{j}+1}^{t-1} s_{l,a_{l}}s_{l,a_{l}}^{\intercal}, C\right) \leq \delta_{t} + \varepsilon\leq 2\varepsilon<1$ (because $\varepsilon< 1/2$). But $\text{Comp}(a,b) \in \{0, 1\}$ for any $a,b\in [0,1]$
                            which means that $\text{Comp}\left(\text{Tr}\left(\text{Dec}_{\textbf{sk}}(\bar{A}_{j})\sum_{l=t_{j}+1}^{t-1} s_{l,a_{l}}s_{l,a_{l}}^{\intercal}\right), C\right) = 0$. But we assumed that
                            $\left|\text{Tr}\left(\text{Dec}_{\textbf{sk}}(\bar{A}_{j})\sum_{l=t_{j}+1}^{t-1} s_{l,a_{l}}s_{l,a_{l}}^{\intercal} \right) - C\right| > \varepsilon'L^{2}\left( \frac{1}{\lambda} + \frac{1}{\sqrt{\lambda}}\right)(t-1-t_{j})$,
                            in other words:
                            \begin{equation}
                                \begin{aligned}
                                &\text{Tr}\left(\text{Dec}_{\textbf{sk}}(\bar{A}_{j})\sum_{l=t_{j}+1}^{t-1} s_{l,a_{l}}s_{l,a_{l}}^{\intercal} \right) > C + \varepsilon'L^{2}\left( \frac{1}{\lambda} + \frac{1}{\sqrt{\lambda}}\right)(t-1-t_{j})\geq C \text{ or } \\
                                &\text{Tr}\left(\text{Dec}_{\textbf{sk}}(\bar{A}_{j})\sum_{l=t_{j}+1}^{t-1} s_{l,a_{l}}s_{l,a_{l}}^{\intercal} \right) < C - \varepsilon'L^{2}\left( \frac{1}{\lambda} + \frac{1}{\sqrt{\lambda}}\right)(t-1-t_{j})
                                \end{aligned}
                            \end{equation}
                            But $\text{Comp}\left(\text{Tr}\left(\text{Dec}_{\textbf{sk}}(\bar{A}_{j})\sum_{l=t_{j}+1}^{t-1} s_{l,a_{l}}s_{l,a_{l}}^{\intercal}\right), C\right) = 0$, it is thus only possible that

                                $$\text{Tr}\left(\text{Dec}_{\textbf{sk}}(\bar{A}_{j})\sum_{l=t_{j}+1}^{t-1} s_{l,a_{l}}s_{l,a_{l}}^{\intercal}\right) \leq C - \varepsilon'L^{2}\left( \frac{1}{\lambda} + \frac{1}{\sqrt{\lambda}}\right)(t-1-t_{j})$$

                        \item If $\left|\text{Tr}\left(\text{Dec}_{\textbf{sk}}(\bar{A}_{j})\sum_{l=t_{j}+1}^{t-1} s_{l,a_{l}}s_{l,a_{l}}^{\intercal} \right) - C\right| \leq \varepsilon'$:

                            In this case, by definition we have
                                {\small$$ \text{Tr}\left(\text{Dec}_{\textbf{sk}}(\bar{A}_{j})\sum_{l=t_{j}+1}^{t-1} s_{l,a_{l}}s_{l,a_{l}}^{\intercal} \right) \leq C + \varepsilon'L^{2}\left( \frac{1}{\lambda} + \frac{1}{\sqrt{\lambda}}\right)(t-1-t_{j})$$}
                    \end{itemize}
            \end{proof}

        The previous proposition ensures that when a batch is ended because $\delta_{t} > 0.45$ then we have, for a small enough $\varepsilon'$, that,
        $\text{Tr}\left(\text{Dec}_{\textbf{sk}}(\bar{A}_{j})\sum_{l=t_{j}+1}^{t-1} x_{l,a_{l}}x_{l,a_{l}}^{\intercal}\right) \geq C'$ for some constant $C'$. However, thanks to Prop.~\ref{prop:convergence_speed_approx_inv}
        we have that for any batch $j$, that for all $l\in \{t_{j}+1, \dots, t-1\}$:
        \begin{align}
            \| x\|_{\bar{V}_{j}^{-1}}^{2} - \| x\|_{\text{Dec}_{\textbf{sk}}(\bar{A}_{j})}^{2} &\leq L^{2}\|\bar{V}_{j}^{-1} - \text{Dec}_{\textbf{sk}}(\bar{A}_{j})\| \leq \frac{L}{t_{j}^{3/2}\sqrt{\lambda + L^{2}t_{j}}}
        \end{align}
        Summing over all time steps $l\in [t_{j}+1, t-1]$, we have that:
        \begin{align}\label{eq:error_trace}
           \left| \text{Tr}\left(\text{Dec}_{\textbf{sk}}(\bar{A}_{j})\sum_{l=t_{j}+1}^{t-1} s_{l,a_{l}}s_{l,a_{l}}^{\intercal}\right) - \text{Tr}\left(\bar{V}_{j}^{-1}\sum_{l=t_{j}+1}^{t-1} s_{l,a_{l}}s_{l,a_{l}}^{\intercal}\right)\right|
           \leq \frac{L(t - 1 - t_{j})}{t_{j}^{3/2}\sqrt{\lambda + L^{2}t_{j}}}
        \end{align}

        Therefore when $\delta_{t} > 0.45$, we that:
        \begin{equation}
            \text{Tr}\left(\bar{V}_{j}^{-1}\sum_{l=t_{j}+1}^{t-1} s_{l,a_{l}}s_{l,a_{l}}^{\intercal}\right) \geq C - \varepsilon_{t}' L^{2}\left( \frac{1}{\lambda} + \frac{1}{\sqrt{\lambda}}\right)(t-1-t_{j}) - \frac{L(t-1-t_{j})}{t_{j}^{3/2}\sqrt{\lambda + L^{2}t_{j}}}
        \end{equation}
        but $\varepsilon_{t}' = \frac{1}{4tL^{2}\left( \frac{1}{\lambda} + \frac{1}{\sqrt{\lambda}}\right)(t-1-t_{j})}$ so:
        \begin{equation}
            \text{Tr}\left(\bar{V}_{j}^{-1}\sum_{l=t_{j}+1}^{t-1} s_{l,a_{l}}s_{l,a_{l}}^{\intercal}\right) \geq C - \frac{1}{4t} - \frac{L(t-1-t_{j})}{t_{j}^{3/2}\sqrt{\lambda + L^{2}t_{j}}}
        \end{equation}
        But if $\delta_{t} \leq 0.45$:
        \begin{equation}
            \text{Tr}\left(\bar{V}_{j}^{-1}\sum_{l=t_{j}+1}^{t-1} s_{l,a_{l}}s_{l,a_{l}}^{\intercal}\right) \leq C + \varepsilon_{t}' L^{2}\left( \frac{1}{\lambda} + \frac{1}{\sqrt{\lambda}}\right)(t-1-t_{j}) + \frac{L(t-1-t_{j})}{t_{j}^{3/2}\sqrt{\lambda + L^{2}t_{j}}}
        \end{equation}
        or using the definition of $\varepsilon_{t}'$:
        \begin{equation}
            \text{Tr}\left(\bar{V}_{j}^{-1}\sum_{l=t_{j}+1}^{t-1} s_{l,a_{l}}s_{l,a_{l}}^{\intercal}\right) \leq C + \frac{1}{4t} + \frac{L(t-1-t_{j})}{t_{j}^{3/2}\sqrt{\lambda + L^{2}t_{j}}}
        \end{equation}

        \subsubsection{Masking Procedure:}\label{app:masking_procedure}

            In order to prevent any leakage of information when the user decrypts the result of this approximate comparison,
            we use a masking procedure where the algorithm adds a big noise to the bit encrypting the approximation of the comparison, somehow masking
            its value to the user. In order for this procedure to be secure, the algorithm needs to sample the noise from a distribution such
            the resulting distribution of the result observed by the user is independent of the value of $\delta_{t}$ (see Prop.~\ref{prop:precision_comparaison_trace}).
            Formally, we add a noise $\xi\sim \Xi$ such that for any $x,x'\in[0,1]$:
            \begin{align}
                \mathbb{P}\left( Dec_{\textbf{sk}}(\xi + x)\right) = \mathbb{P}\left( Dec_{\textbf{pk}}(\xi + x')\right)
            \end{align}

            Finding such distribution is highly dependent on the encryption scheme used and its parameters. In our implementation,
            we used the CKKS scheme with depth $D = 100$, level of security $\kappa = 128$ and a log size of modulus $\log_{2}(q_{0}) = 4982$.
            Therefore, for a cyphertext $ct$ at a given level $l$ encrypting a number $x\in[0,1]$ with a pair of public and secret key $(\textbf{pk}, \textbf{sk})$ we have that:
            \begin{align}
                Dec_{\textbf{sk}}(ct) = \langle ct, \textbf{sk} \rangle (\text{mod} q_{l})
            \end{align}
            with $q_{l} = 2^{l}q_{0}$. When sampling an integer $r$ uniformly in $\{0, \dots, q_{l}-1\}$, we have that for any $k\in\{0, \dots, q_{l}-1\}$ the distribution of $r + k (\text{mod} q_{l})$ is uniform
            over $\{0, \dots, q_{l}-1\}$. We leverage this result to creates a masking procedure detailed in Alg.~\ref{alg:masking_procedure}

            \begin{algorithm}[H]
                \small
                \caption{Masking Procedure}
                \label{alg:masking_procedure}
                \begin{algorithmic}
                    \STATE {\bfseries Input:} ciphertext: $ct$, modulus factor: $q_{l}$, cyclotomial polynomial degree: $M$
                    \STATE Sample uniformly $r$ in $\{0, \dots, q_{l}-1\}$
                    \STATE Compute the polynomial $\wt{r} \in \mathbb{Z}[X]/(X^{M} + 1)$ such that $\wt{r}(X) = r$
                    \STATE {\bfseries Return:} $ct + \wt{r}$
                \end{algorithmic}
            \end{algorithm}

            Upon receiving the decryption of $ct + \wt{r}$, the unmasking procedure is consists in simply subtracting $r$.

        \subsubsection{Impact on the Growth of the determinant:}

            In this section, we study the impact on the determinant  of the design matrix when Condition~\ref{eq:trace_condition} is satisfied
            for some constant $C'$. Our result is based on the classic following lemma.

            \begin{lemma}\label{lemma:bound_det}
				For any positive definite symmetric matrix $A, B$ and symmetric semi-positive definite matrix $C$ such that $A = B + C$ we have that:
				\begin{align}
					\frac{\text{det}(A)}{\text{det}(B)} \geq 1 + \text{Tr}\left(B^{-1/2}CB^{-1/2} \right)
				\end{align}
			\end{lemma}

            \begin{proof}{of Lemma~\ref{lemma:bound_det}.} Using that $A = B + C$:
				\begin{align}
					\frac{\text{det}(A)}{\text{det}(B)} = \text{det}\left( I_{d} +B^{-1/2}CB^{-1/2}\right) \geq 1 + \text{Tr}\left(B^{-1/2}CB^{-1/2} \right) = 1 + \text{Tr}\left(B^{-1}C \right)
				\end{align}
                The last inequality is a consequence of the following inequality:
                \begin{align}
                    \forall n\in\mathbb{N}^{\star}, \forall a\in \mathbb{R}^{n}_{+}, \qquad 1 + \sum_{i=1}^{n} a_{i} \leq \prod_{i=1}^{n} ( 1 + a_{i})
                \end{align}
                Indeed $I_{d} + B^{-1/2}CB^{-1/2}$ is symmetric definite positive hence its   eignevalues are positive.
            \end{proof}

            Therefore, using the lemma above applied to the design matrix, $V_{t} = \bar{V}_{j} + \sum_{l=t_{j} + 1}^{t-1} s_{l,a_l}s_{l,a_l}^{\intercal}$ for $t\geq t_{j} + 1$, we have that:
            \begin{equation}
                \text{det}\left( V_{t} \right) \geq \left( 1 + \text{Tr}\left(\bar{V}_{j}^{-1} \sum_{l=t_{j} + 1}^{t-1} s_{l,a_l}s_{l,a_l}^{\intercal}\right)\right)\text{det}(\bar{V}_j)
            \end{equation}

        \subsubsection{Putting Everything Together:}

            We are finally, ready to prove an upper-bound on the number of batches. First, let's recall Prop.~\ref{prop:nb_batches}.

            \begin{proposition*}
                If $ C - \frac{L\eta}{\sqrt{\lambda + L^{2}}} > \frac{1}{4}$, the number of episodes in Alg.~\ref{alg:low_switching_algo}, $M_{T}$ for $T$ steps, is bounded by:
                \begin{align}
                    M_{T} \leq 1 + \frac{d\ln\left(1 + \frac{L^{2}T}{\lambda d}\right)}{2\ln\left( \frac{3}{4} + C - \frac{L\eta}{\sqrt{\lambda + L^{2}}} \right)} + \frac{\ln(T)}{\ln(1 + \eta)}
                \end{align}
            \end{proposition*}
            \begin{proof}{of Lem.~\ref{prop:nb_batches}.} Let's define for $i\geq 1$, the macro-episode:
                \begin{align}
                    n_{i} = \min\left\{t> n_{i-1}\mid \delta_{t} > \varepsilon_{t}\right\}
                \end{align}
                with $n_{0} = 0$. In other words, macro-episodes are episodes such that the norm of the context has grown too big. It means that for all episodes between two
                macro-episodes the batches are ended because the current batch is too long. Therefore for macro-episode $i$, thanks to Eq.~\eqref{eq:error_trace} and Prop.~\ref{prop:precision_comparaison_trace}:
                \begin{align}
                    C - \varepsilon_{n_{i}}'L^{2}\left( \frac{1}{\lambda} + \frac{1}{\sqrt{\lambda}}\right)(n_{i}-1-t_{j}) - \frac{L(n_{i}-1 - t_{j})}{t_{j}^{3/2}\sqrt{\lambda + L^{2}t_{j}}}
                    \leq \text{Tr}\left(\bar{V}_{j}^{-1}\sum_{l=t_{j}+1}^{n_{i}-1} s_{l,a_{l}}s_{l,a_{l}}^{\intercal} \right)
                \end{align}
                where $\varepsilon_{n_{i}}' = \left(4n_{i}L^{2}\left( \frac{1}{\lambda} + \frac{1}{\sqrt{\lambda}}\right)(n_{i}-1-t_{j})\right)^{-1}$ as defined in Alg.~\ref{alg:low_switching_algo}.
                But the batch $j$ for which $n_{i} = t_{j+1}$ is such that $t_{j+1} - t_{j} \leq \eta t_{j} +1$ (thanks to the second $\text{if}$ condition in Alg.~\ref{alg:low_switching_algo}). Hence:
                \begin{align}
                    \frac{L(n_{i}-1 - t_{j})}{t_{j}^{3/2}\sqrt{\lambda + L^{2}t_{j}}} \leq \frac{L\eta }{\sqrt{t_{j}(\lambda + L^{2}t_{j})}}\leq \frac{L\eta}{\sqrt{\lambda + L^{2}}}
                \end{align}
                Therefore by Lem.~\ref{lemma:bound_det} we have that:
                \begin{align}
                    \det{\bar{V}_{j+1}} \geq \left(1 + C - \left(\frac{1}{4} + \frac{L\eta}{\sqrt{\lambda + L^{2}}}\right)\right)\det{\bar{V}_{j}}
                \end{align}
                because for all $t$, $\varepsilon_{t}' L^{2}\left( \frac{1}{\lambda} + \frac{1}{\sqrt{\lambda}}\right)(t-1-t_{j})\leq \frac{1}{4}$ and because for any episode $j$ between two macro-episodes $i$ and $i+1$ the determinant
                of the design matrix is an increasing function of the episode (because for two matrices $M,N$ symmetric semi-definite positive $\text{det}(M+N) \geq \text{det}(M)$).

                Thus $\det{\bar{V}_{j_{i}}} \geq (\frac{3}{4} + C - \frac{L\eta}{\sqrt{\lambda + L^{2}}})\det{\bar{V}_{j_{i-1}}}$ where $j_{i}$ is the episode such that $n_{i} = t_{j_{i}+1}$. Therefore
                the number of macro-episodes $M_{1}$ is such that:
                \begin{align}
                    \left(\frac{3}{4} + C - \frac{L\eta}{\sqrt{\lambda + L^{2}}}\right)^{M_{1}-1} \leq  \frac{\text{det}(\bar{V}_{M_{T}})}{\text{det}(\bar{V}_{0})}
                \end{align}
                where $\bar{V}_{M_{T}}$ is the design matrix after $T$ steps (or $M_{T}$ batches) and $\bar{V}_{0} = \lambda I_{d}$. This upper bound gives that:
                \begin{align}
                    M_{1} \leq 1 + \frac{\ln\left(\frac{\text{det}(\bar{V}_{M_{T}})}{\text{det}(\bar{V}_{0})}\right)}{\ln\left( \frac{3}{4} + C - \frac{L\eta}{\sqrt{\lambda + L^{2}}} \right)} 
                \end{align}
                if $\frac{3}{4} + C - \frac{L\eta}{\sqrt{\lambda + L^{2}}} > 1$. Moreover, thanks to Lemma $10$ in \cite{abbasi2011improved}, the
                log-determinant of the design matrix is bounded by:
                $\ln\left(\frac{\text{det}(\bar{V}_{M_{T}})}{\text{det}(\bar{V}_{0})}\right) \leq d\ln\left(1 + \frac{TL^{2}}{\lambda d}\right)$.
                In addition, there is at most $1 + \frac{\ln(n_{i+1}/n_{i})}{\ln(1 + \eta)}$ batches between macro-episode $i$ and $i+1$. Therefore:
                \begin{align*}
                    M_{T} \leq \sum_{i=0}^{M_{1}-1} 1 + \frac{1}{\ln(1 + \eta)}\ln(n_{i+1}/n_{i}) &= M_{1} + \frac{\ln(T)}{\ln(1 + \eta)}
                    \\&\leq 1 + \frac{d\ln\left(1 + \frac{L^{2}T}{\lambda d}\right)}{2\ln\left( \frac{3}{4} + C - \frac{L\eta}{\sqrt{\lambda + L^{2}}} \right)} + \frac{\ln(T)}{\ln(1 + \eta)}
                \end{align*}

            \end{proof}

    \subsection{Regret Upper Bound (Proof of Thm.~\ref{thm:regret_efficient_linucb})}\label{app:proof_regret_efficient_linucb}

        Now that we have shown an upper-bound on the number of bathes for the \secofulls algorithm, we are ready to prove the regret bound of
        Thm.~\ref{thm:regret_efficient_linucb}. The proof of this theorem follows the same logic as the regret analysis of \oful.
        That is to say, we first show a high-probability upper bound on the regret thanks to optimism and then proceed to bound each term
        of the bonus used in \secofulls.

        We first show the following lemma giving a first upper bound on the regret relating the error due to the approximation
        of the argmax and optimism.
        \begin{lemma}\label{lem:first_bound_regret}
            For any $\delta>0$, the regret of Alg.~\ref{alg:low_switching_algo} is bounded with probability at least $1- \delta$ by:
            {\small\begin{equation}\label{eq:first_bound_regret}
                \begin{aligned}
                     R_{T}(\secofulls)\leq
                     \underbrace{\sum_{j=0}^{M_{T}-1} \sum_{t=t_{j}+1}^{t_{j+1}}\frac{4}{t} \left(1 + 2\wt{\beta}(j)\left[\frac{2}{t}
                     + \sqrt{\frac{L}{t_{j}^{3/2}\sqrt{t_{j}L^{2}+ \lambda}}} + L\sqrt{\frac{1}{\lambda} +
                     \frac{1}{\sqrt{\lambda}}} \right] \right)}_{:= \textcircled{1}} \\
                     +\underbrace{\sum_{j=0}^{M_{T}-1} \sum_{t= t_{j}+1}^{t_{j+1}} 2\wt{\beta}(j)\left[ \mathrm{sqrt}_{\mathrm{HE}}\left( s_{t,a}^\top \text{Dec}_{\textbf{sk}}(\bar{A}_{j}) s_{t,a} + \frac{L}{t_{j}^{3/2}\sqrt{\lambda + L^{2}t_{j}}}\right) + \frac{1}{t}\right]}_{:= \textcircled{2}}
                \end{aligned}
            \end{equation}}
            where for every time step $t$, $a_{t}^{\star} = \arg\max_{a\in [K]} \langle x_{t,a}, \theta^{\star} \rangle$,
            $M_{T}$ is the number of batches and $ R_{T}(\secofulls) = \sum_{t=1}^{T} \langle \theta^{\star}, s_{t,a_{t}^{\star}} - s_{t,a_{t}} \rangle$. 
        \end{lemma}

        \begin{proof}{of Lem.~\ref{lem:first_bound_regret}.}
            First, let's define $E$ the event that all confidence ellipsoids, $\tilde{\mathcal{C}}_{j}$, contain $\theta^\star$
            with probability at least $1 - \delta$. That is to say $E = \left\{\theta^{\star} \in \bigcap_{j=1}^{+\infty} \tilde{\mathcal{C}}_{j}(\delta)\right\}$.
            Thanks to Prop.~\ref{prop:confidence_theta}, $\mathbb{P}\left(E\right) \geq 1 - \delta$. Because $E$ is included in
            the event described by Prop.~\ref{prop:confidence_theta}.

            Therefore conditioned on the event $E$, after $T$ steps the regret can be decomposed as:
            \begin{equation}
                \begin{aligned}
                R_{T}(\text{\secofulls}) &= \sum_{t=1}^{T} \langle \theta^{\star}, s_{t,a_{t}^{\star}} \rangle - \max_{a\leq K} \text{Dec}_{\textbf{sk}}(\rho_{a}(t)) + \max_{a\leq K}\text{Dec}_{\textbf{sk}}(\rho_{a}(t)) - \text{Dec}_{\textbf{sk}}(\rho_{a_{t}}(t)) \\
                &+ \text{Dec}_{\textbf{sk}}(\rho_{a_{t}}(t)) - \langle \theta^{\star}, s_{t,a_{t}} \rangle
                \end{aligned}
            \end{equation}
            where $\rho_{a}(t)$ is the optimistic upper bound on the reward of arm $a$ computed by Alg.~\ref{alg:low_switching_algo} and $a_{t}^{\star} = \arg\max_{a\in [K]} \langle \theta^{\star}, s_{t,a_{t}^{\star}}\rangle$.
            Now for any $t\leq T$, under the event $E$,
            $\langle \theta^{\star}, s_{t,a_{t}^{\star}} \rangle \leq \max_{a\leq K} \text{Dec}_{\textbf{sk}}(\rho_{a}(t))$. But thanks to Cor.~\ref{cor:precision_argmax}, for any $t\geq 1$ inside batch $j$:
            {\small\begin{align}
                \max_{a} \text{Dec}_{\textbf{sk}}(\rho_{a}(t)) - \text{Dec}_{\textbf{sk}}(\rho_{a_{t}}(t)) \leq \frac{1}{t}\left(1 + \wt{\beta}(j)\left[\frac{2}{t} +  \sqrt{\frac{L}{t_{j}^{3/2}\sqrt{\lambda + L^{2}t_{j}}}} + L\sqrt{\frac{1}{\lambda} + \frac{1}{\sqrt{\lambda}}}\right]\right)
            \end{align}}

            In addition, we have that under event $E$:
            {\small\begin{equation*}
                \text{Dec}_{\textbf{sk}}(\rho_{a_{t}}(t)) - \langle \theta^{\star}, s_{t,a_{t}}\rangle = \langle \tilde{\theta}_{j} - \theta^{\star}, s_{t,a_{t}} \rangle + \wt{\beta}(j)\left[\mathrm{sqrt}_{\mathrm{HE}}\left( s_{t,a}^\top \text{Dec}_{\textbf{sk}}(\bar{A}_{j}) s_{t,a} + \frac{L}{t_{j}^{3/2}\sqrt{\lambda + L^{2}t_{j}}}\right) + \frac{1}{t} \right]
            \end{equation*}}
            But still conditioned on the event $E$, $\langle \tilde{\theta}_{j} - \theta^{\star}, s_{t,a_{t}} \rangle \leq \|\tilde{\theta}_{j} - \theta^{\star}\|_{\bar{V}_{j}}\| s_{t,a_{t}}\|_{\bar{V}_{j}^{-1}} \leq \tilde{\beta}(j)\| s_{t,a_{t}}\|_{\bar{V}_{j}^{-1}}
            \leq \wt{\beta}(j)\left[ \mathrm{sqrt}_{\mathrm{HE}}\left( s_{t,a_{t}}^\top \text{Dec}_{\textbf{sk}}(\bar{A}_{j}) s_{t,a_{t}} + \frac{L}{t_{j}^{3/2}\sqrt{\lambda + L^{2}t_{j}}}\right) + \frac{1}{t}\right]$.
            Putting the last two equations together, for every step $t\leq T$:
            \begin{align*}
                \langle \theta^{\star}, s_{t,a_{t}^{\star}} - s_{t,a_{t}} \rangle \leq &\frac{1}{t}\left(1 + \wt{\beta}(j)\left[\frac{2}{t} +  \sqrt{\frac{L}{t_{j}^{3/2}\sqrt{\lambda + L^{2}t_{j}}}} + L\sqrt{\frac{1}{\lambda} + \frac{1}{\sqrt{\lambda}}}\right]\right) \\
                &+ 2\wt{\beta}(j)\left[\mathrm{sqrt}_{\mathrm{HE}}\left( s_{t,a_{t}}^\top \text{Dec}_{\textbf{sk}}(\bar{A}_{j}) s_{t,a_{t}} + \frac{L}{t_{j}^{3/2}\sqrt{\lambda + L^{2}t_{j}}}\right)+ \frac{1}{t} \right]
            \end{align*}
        \end{proof}

        \paragraph{Bounding \textcircled{1}.} We now proceed to bound each term in Eq.~\eqref{eq:first_bound_regret}. The following
        lemma is used to \textcircled{1}.
            \begin{lemma}\label{lem:bound_error_argmax}
                For all $t\geq 1$:
                \begin{align}
                    \sum_{j=0}^{M_{T}-1} \sum_{t=t_{j}+1}^{t_{j+1}}\frac{1}{t} \left(1 + \wt{\beta}(j)\left[\frac{2}{t}
                + \sqrt{\frac{L}{t_{j}^{3/2}\sqrt{t_{j}L^{2}+ \lambda}}} + L\sqrt{\frac{1}{\lambda} +
                \frac{1}{\sqrt{\lambda}}} \right] \right) \leq \mathcal{O}(\ln(T)^{3/2})
                \end{align}
            \end{lemma}

            \begin{proof}{of Lem.~\ref{lem:bound_error_argmax}.} Because $t_j \geq 1$:
				\begin{equation}
					\frac{L}{t_{j}^{3/2}\sqrt{\lambda + L^{2}t_{j}}} \leq \frac{L}{\sqrt{\lambda}},~~~\wt{\beta}(j) \leq 1 + \sqrt{\lambda}S + \sigma\sqrt{d\left(\ln\left(1 + \frac{L^{2}T}{\lambda d}\right) +  \ln\left(\frac{\pi^{2}T^{2}}{6\delta}\right)\right)}
				\end{equation}
				Bounding each component of the sum of \textcircled{1} in Eq.~\eqref{eq:first_bound_regret} individually, we get:
				\begin{align}
					\sum_{j=0}^{M_{T}-1} \sum_{t=t_{j}+1}^{t_{j+1}} \frac{1}{t}\leq (1 + \ln(T))
				\end{align}
				Hence:
				\begin{align*}
					\sum_{j=0}^{M_{T}-1} \sum_{t=t_{j}+1}^{t_{j+1}}\frac{1}{t} \left(1 + \wt{\beta}(j)\left[\frac{2}{t}
                    + \sqrt{\frac{L}{t_{j}^{3/2}\sqrt{t_{j}L^{2}+ \lambda}}} + L\sqrt{\frac{1}{\lambda} +
                    \frac{1}{\sqrt{\lambda}}} \right] \right) \leq \left( 1 + \ln(T)\right) \Bigg[1 + &\\
                    \left(1 + \sqrt{\lambda}S + \sigma\sqrt{d\left(\ln\left(1 + \frac{L^{2}T}{\lambda d}\right) +  \ln\left(\frac{\pi^{2}T^{2}}{6\delta}\right)\right)} \right)
                    \left(2 + L\sqrt{\frac{1}{\lambda} + \frac{1}{\sqrt{\lambda}}} + \sqrt{\frac{L}{\sqrt{\lambda}}}\right)\Bigg]&
				\end{align*}
            \end{proof}

            Lem.~\ref{lem:bound_error_argmax} shows that the error from our procedure to select the argmax induces only an additional
            logarithmic cost in $T$ compared with the regret of directly selecting the argmax of the UCBs $(\rho_{a}(t))_{a\leq K}$.

        \paragraph{Bounding \textcircled{2}.}

            We are now left with bounding the second term in Eq.~\eqref{eq:first_bound_regret}. This term is usually the one that
            appears in regret analysis for linear contextual bandits. First, \textcircled{2} can be further broke down thanks to the following lemma.

            \begin{lemma}\label{lem:second_bound_regret}
				For all $t\geq 1$,
				{\small\begin{equation}
					\begin{aligned}
                        \sum_{j=0}^{M_{T}-1} \sum_{t=t_{j}+1}^{t_{j+1}} \wt{\beta}(j)\left[ \mathrm{sqrt}_{\mathrm{HE}}\left( s_{t,a_{t}}^\top \text{Dec}_{\textbf{sk}}(\bar{A}_{j}) s_{t,a_{t}} + \frac{L}{t_{j}^{3/2}\sqrt{\lambda + L^{2}t_{j}}}\right) + \frac{1}{t}\right]
                        \leq \underbrace{\sum_{j=0}^{M_{T}-1} \sum_{t=t_{j}+1}^{t_{j+1}} \frac{2\wt{\beta}(j)}{t}}_{ := \textcircled{a}}&\\
                             +\underbrace{\sum_{j=0}^{M_{T}-1} \sum_{t=t_{j}+1}^{t_{j+1}} \wt{\beta}(j)\|s_{t,a_{t}}\|_{\bar{V}_{j}^{-1}}}_{:= \textcircled{b}} +\underbrace{\sum_{j=0}^{M_{T}-1} \sum_{t=t_{j}+1}^{t_{j+1}} \wt{\beta}(j)\sqrt{\frac{2L}{t_{j}^{3/2}\sqrt{\lambda + L^{2}t_{j}}}}}_{ := \textcircled{c}}&
					\end{aligned}
				\end{equation}}
			\end{lemma}

			\begin{proof}{of Lem.~\ref{lem:second_bound_regret}.}
				For any time $t\geq 1$ thanks to Prop.~\ref{prop:iterations_square_root}, we have:
				{\small\begin{align*}
					\mathrm{sqrt}_{\mathrm{HE}}\left( s_{t,a_{t}}^\top \text{Dec}_{\textbf{sk}}(\bar{A}_{j}) s_{t,a_{t}} + \frac{L}{t_{j}^{3/2}\sqrt{\lambda + L^{2}t_{j}}}\right) \leq  \frac{1}{t} + \sqrt{\| s_{t,a_{t}}\|_{\text{Dec}_{\textbf{sk}}(\bar{A}_{j})}^{2} + \frac{L}{t_{j}^{3/2}\sqrt{\lambda + L^{2}t_{j}}}}&\\
					\leq \frac{1}{t} + \sqrt{\frac{L}{t_{j}^{3/2}\sqrt{\lambda + L^{2}t_{j}}} + \|s_{t,a_{t}}\|_{\bar{V}_{j}^{-1}}^{2} + \|s_{t,a_{t}}\|_{2}^{2}\|\text{Dec}_{\textbf{sk}}(\bar{A}_{j})- \bar{V}_{j}^{-1}\|}&\\
					\leq \frac{1}{t} + \sqrt{\frac{2L}{t_{j}^{3/2}\sqrt{\lambda + L^{2}t_{j}}} + \|s_{t,a_{t}}\|_{\bar{V}_{j}^{-1}}^{2}}&\\
					\leq \frac{1}{t} + \sqrt{\frac{2L}{t_{j}^{3/2}\sqrt{\lambda + L^{2}t_{j}}}} + \|s_{t,a_{t}}\|_{\bar{V}_{j}^{-1}}&
				\end{align*}}
            \end{proof}

            We proceed to bound each term \textcircled{a}, \textcircled{b}, \textcircled{c}. Bounding \textcircled{b} is similar to the
            analysis of \oful. On the other hand, bounding neatly  \textcircled{c} is the reason why we introduced the condition that a new episode is started is $t\geq (1 + \eta)t_{j}$.

            The following lemma bounds \textcircled{a} which is simply a numerical error due to the approximation of the square root.
			\begin{lemma}\label{lem:bound_precision_sqrt}
				For any $T\geq 1$,
				{\small\begin{align}
					\sum_{j=0}^{M_{T}-1} \sum_{t=t_{j}+1}^{t_{j+1}} \frac{4\wt{\beta}(j)}{t} \leq 4\left(1 + \sqrt{\lambda}S + \sigma\sqrt{d\left(\ln\left(1 + \frac{L^{2}T}{\lambda d}\right) +  \ln\left(\frac{\pi^{2}T^{2}}{6\delta}\right)\right)}\right) ( 1 + \ln(T))
				\end{align}}
			\end{lemma}
			\begin{proof}{of Lem.~\ref{lem:bound_precision_sqrt}.}
				Using the upper bound on the $\wt{\beta}(j)$ shown in the proof of Lem.~\ref{lem:bound_error_argmax}, we get the result.
            \end{proof}

            We are finally left with the two terms \textcircled{b} and \textcircled{c}. The first term, \textcircled{b}, will be compared to the
            bonus used in \oful so that we can use Lemma $11$ in \cite{abbasi2011improved} to bound it. But first, we need
            to show how the norm for two different matrices $A$ and $B$ relates to each other.

            \begin{lemma}\label{lemma:bound_norm_context}
				For any context $x\in\mathbb{R}^{d}$ and symmetric semi-definite matrix $A$,$B$ and $C$ such that $A = B + C$ then:
				\begin{align}
					\|x\|_{B^{-1}}^{2} \leq \lambda_{\max}\left( I_{d} + B^{-1/2}CB^{-1/2}\right) \|x\|_{A^{-1}}^{2} \leq \left(1 + \text{Tr}\left( B^{-1/2}CB^{-1/2}\right)\right) \|x\|_{A^{-1}}^{2}
                \end{align}
                where $\lambda_{\max}(.)$ returns the maximum eigenvalue of a matrix.
			\end{lemma}

            \begin{proof}{of Lemma~\ref{lemma:bound_norm_context}.} We have by definition of $A$ and $B$:
				\begin{align}
					\langle x, A^{-1}x\rangle = \langle x, (B + C)^{-1}x\rangle &= \langle x, B^{-1/2}(I_{d} + B^{-1/2}CB^{-1/2})^{-1}B^{-1/2}x\rangle\\
					&= \langle B^{-1/2}x, (I_{d} + B^{-1/2}CB^{-1/2})^{-1} (B^{-1/2}x)\rangle\\
					&\geq \lambda_{\min}\left((I_{d} + B^{-1/2}CB^{-1/2})^{-1}\right) \|B^{-1/2}x\|^{2}\\
					&\geq \frac{1}{\lambda_{\max}(I_{d} + B^{-1/2}CB^{-1/2})}\|x\|_{B^{-1}}^{2}
				\end{align}
				Hence:
				\begin{align}
					\|x\|_{B^{-1}}^{2} \leq \lambda_{\max}\left( I_{d} + B^{-1/2}CB^{-1/2} \right) \|x\|_{A^{-1}}^{2}
				\end{align}
                The result follows from Weyl's inequality \cite{horn_johnson_1991}, that is to say for all symmetric matrix $M,N$ $\lambda_{\max}(M + N) \leq \lambda_{\max}(M) + \lambda_{\max}(N)$.
                And the fact that all eigenvalues of $B^{-1/2}CB^{-1/2}$ are positive hence $\lambda_{\max}(B^{-1/2}CB^{-1/2}) \leq \text{Tr}(B^{-1/2}CB^{-1/2}) = \text{Tr}(CB^{-1})$.
			\end{proof}
            We are now able to bound \textcircled{b} using Lemma $11$ in \cite{abbasi2011improved}.
            \begin{lemma}\label{lem:bound_sum_norm_context}
				If $\lambda\geq L^{2}$ we have:
				{\small\begin{equation}
					\begin{aligned}
                    \sum_{j=0}^{M_{T}-1} \sum_{t=t_{j}+1}^{t_{j+1}} 2\beta(j)\|s_{t,a_{t}}\|_{\bar{V}_{j}^{-1}} \leq \beta^{\star}\sqrt{2d\ln\left(1 + \frac{TL^{2}}{\lambda d}\right)}\Bigg[\sqrt{T\left(1.25 + C + L^{2}\left( \frac{1}{\lambda} + \frac{1}{\sqrt{\lambda}}\right)\right)} \\
                    + \sqrt{M_{T}\left(\eta^{2} + \frac{L}{(\lambda + L^{2})^{3/2}} \right)}\Bigg]&
					\end{aligned}
				\end{equation}}
				with $\beta^{\star} = 1 + \sqrt{\lambda}S + \sigma\sqrt{d\left(\ln\left(1 + \frac{L^{2}T}{\lambda d}\right) +  \ln\left(\frac{\pi^{2}T^{2}}{6\delta}\right)\right)}$
			\end{lemma}
			\begin{proof}{of Lem.~\ref{lem:bound_sum_norm_context}.} For any time $t$ in batch $j$, we have thanks to Lem.~\ref{lemma:bound_norm_context} that:
				\begin{align}
					\| s_{t,a_{t}} \|_{\bar{V}_{j}^{-1}} \leq \sqrt{1 + \text{Tr}\left(\bar{V}_{j}^{-1} \sum_{l=t_{j}+1}^{t-1} s_{l,a_{l}}s_{l,a_{l}}^{\intercal} \right)} \| s_{t,a_{t}}\|_{V_{t}^{-1}}
				\end{align}
				with $V_{t} = \lambda I_{d} + \sum_{l=1}^{t-1} s_{l,a_{l}}s_{l,a_{l}}^{\intercal}$.
                The rest of the proof relies on bounding $\text{Tr}\left(\bar{V}_{j}^{-1} \sum_{l=t_{j}+1}^{t-1} s_{l,a_{l}}s_{l,a_{l}}^{\intercal} \right)$.
                To do so, we will use the following inequality, see Eq.~\eqref{eq:error_trace}:
				\begin{align*}
                    -\frac{L(t-1 - t_{j})}{t_{j}^{3/2}\sqrt{\lambda + L^{2}t_{j}}} \leq \text{Tr}\left(\left(\bar{V}_{j}^{-1} - \text{Dec}_{\textbf{sk}}(\bar{A}_{j})\right)\sum_{l=t_{j}+1}^{t-1} s_{l,a_{l}}s_{l,a_{l}}^{\intercal} \right)
                    \leq \frac{L(t-1 - t_{j})}{t_{j}^{3/2}\sqrt{\lambda + L^{2}t_{j}}}&
				\end{align*}
                Therefore$ \delta_{t}\leq 0.45$  during batch $j$, because  it is not over while no condition is satisfied.
                Thanks to Prop.~\ref{prop:precision_comparaison_trace} with $\varepsilon' = \frac{1}{4tL^{2}(t-1-t_{j})}$, we get:
                    $$\forall t\in \{t_{j}+1,\dots t_{j+1}-1\}, \qquad\text{Tr}\left(\text{Dec}_{\textbf{sk}}(\bar{A}_{j})\sum_{l=t_{j}+1}^{t-1} s_{l,a_{l}}s_{l,a_{l}}^{\intercal} \right) \leq C + \frac{t-1-t_{j}}{4t}$$
                However, for $t = t_{j+1}$ we have either that $\delta_{t_{j+1}} > 0.45$ or $t_{j+1} \geq (1+ \eta)t_{j}$:
                \begin{itemize}
                    \item If $\delta_{t_{j+1}} \leq 0.45$, then $\text{Tr}\left(\text{Dec}_{\textbf{sk}}(\bar{A}_{j})\sum_{l=t_{j}+1}^{t_{j+1}-1} s_{l,a_{l}}s_{l,a_{l}}^{\intercal} \right) \leq C + \frac{t_{j+1}-1-t_{j}}{4t}$
                    \item If $\delta_{t_{j+1}} > 0.45$, then $\text{Tr}\left(\text{Dec}_{\textbf{sk}}(\bar{A}_{j})\sum_{l=t_{j}+1}^{t_{j+1}-1} s_{l,a_{l}}s_{l,a_{l}}^{\intercal} \right) \geq C - \frac{t_{j+1}-1-t_{j}}{4t}$ but $\delta_{t_{j+1}-1} \leq 0.45$ thus
                    $\text{Tr}\left(\text{Dec}_{\textbf{sk}}(\bar{A}_{j})\sum_{l=t_{j}+1}^{t_{j+1}-2} s_{l,a_{l}}s_{l,a_{l}}^{\intercal} \right) \leq C + \frac{t-1-t_{j}}{4t}$. Therefore, using that $\| s_{t,a_{t}}\|_{\text{Dec}_{\textbf{sk}}(\bar{A}_{j})}^{2} \leq \lambda_{\max}(\text{Dec}_{\textbf{sk}}(\bar{A}_{j}))\|s_{t,a_{t}}\|_{2}^{2} \leq L^{2}\left( \frac{1}{\lambda} + \frac{1}{L\sqrt{\lambda + L^{2}}}\right)$.
                    Hence, we have that:
                    \begin{align}
                        \text{Tr}\left(\text{Dec}_{\textbf{sk}}(\bar{A}_{j})\sum_{l=t_{j}+1}^{t_{j+1}-1} s_{l,a_{l}}s_{l,a_{l}}^{\intercal} \right) \leq C + \frac{t-1-t_{j}}{4t} + L^{2}\left(\frac{1}{\lambda} + \frac{1}{L\sqrt{\lambda + L^{2}}}\right)
                    \end{align}
                \end{itemize}

                To sum up, for all $t_{j}+1\leq t\leq t_{j+1}$:
				\begin{align}
					\text{Tr}\left(\bar{V}_{j}^{-1} \sum_{l=t_{j}+1}^{t-1} s_{l,a_{l}}s_{l,a_{l}}^{\intercal}\right)  \leq C + \frac{t-1-t_{j}}{4t} + L^{2}\left(\frac{1}{\lambda} + \frac{1}{\sqrt{\lambda}}\right)  + \frac{L(t-1 - t_{j})}{t_{j}^{3/2}\sqrt{\lambda + L^{2}t_{j}}}
                \end{align}
				Overall, we have that:
				{\small \begin{align}
                    &\sum_{j=0}^{M_{T}-1} \sum_{t=t_{j}+1}^{t_{j+1}} \| s_{t,a_{t}}\|_{\bar{V}_{j}^{-1}} \leq \sum_{j=0}^{M_{T}-1} \sum_{t=t_{j}+1}^{t_{j+1}} \sqrt{1 + \text{Tr}\left(\bar{V}_{j}^{-1} \sum_{l=t_{j}+1}^{t-1} s_{l,a_{l}}s_{l,a_{l}}^{\intercal} \right)}\|s_{t,a_{t}}\|_{V_{t}^{-1}}\\
					&\leq  \sum_{j=0}^{M_{T}-1} \sum_{t=t_{j}+1}^{t_{j+1}} \| s_{t,a_{t}}\|_{V_{t}^{-1}} \sqrt{1 +C + \frac{t-1-t_{j}}{4t} + L^{2}\left(\frac{1}{\lambda} + \frac{1}{\sqrt{\lambda}}\right)  + \frac{L(t-1 - t_{j})}{t_{j}^{3/2}\sqrt{\lambda + L^{2}t_{j}}}}\\
					&\leq  \sqrt{\frac{5}{4} + C + L^{2}\left( \frac{1}{\lambda} + \frac{1}{\sqrt{\lambda}}\right)}\sqrt{T\sum_{t=1}^{T} \|s_{t,a_{t}}\|_{V_{t}^{-1}}^{2}} + \sum_{j=0}^{M_{T}-1} \sum_{t=t_{j}+1}^{t_{j+1}} \| s_{t,a_{t}}\|_{V_{t}^{-1}} \sqrt{\frac{L(t-1 - t_{j})}{t_{j}^{3/2}\sqrt{\lambda + L^{2}t_{j}}}} \label{eq:last_bound_oful}
				\end{align}}
				where the last inequality is due to Cauchy-Schwarz inequality. The first term in inequality Eq.~\eqref{eq:last_bound_oful} is bounded by using Lemma $29$ in \cite{ruan2020linear},
				\begin{align}
					\sum_{t=1}^{T} \|x_{t,a_{t}}\|_{V_{t}^{-1}}^{2} \leq 2\ln\left( \frac{\text{det}(V_{T})}{\text{det}(V_{0})}\right) \leq 2d\ln\left(1 + \frac{TL^{2}}{\lambda d} \right)
				\end{align}
				In addition, the last term in Eq.~\eqref{eq:last_bound_oful} is bounded by:
				\begin{align}
					\sum_{j=0}^{M_{T}-1} \sum_{t=t_{j}+1}^{t_{j+1}} \frac{L(t-1-t_{j})}{t_{j}^{3/2}\sqrt{\lambda + L^{2}t_{j}}} \leq \sum_{j=0}^{M_{T}-1} \frac{L(t_{j+1} - t_{j})^{2}}{2t_{j}^{3/2}\sqrt{\lambda + L^{2}t_{j}}}
				\end{align}
				But a consequence of the second condition is that the length of batch $j$ satisfies $t_{j+1} - t_{j} \leq \eta t_{j} + 1$. Therefore:
				\begin{align}
					\sum_{j=0}^{M_{T}-1} \frac{Lf(t_{j+1} - t_{j})^{2}}{2t_{j}^{3/2}\sqrt{\lambda + L^{2}t_{j}}} &\leq \sum_{j=0}^{M_{T}-1} \frac{L(\eta t_{j} + 1)^{2}}{2t_{j}^{3/2}\sqrt{\lambda + L^{2}t_{j}}} \\
					&\leq \sum_{j=0}^{M_{T}-1} \frac{L(\eta^{2}t_{j}^{2} +1)}{t_{j}^{3/2}\sqrt{\lambda + L^{2}t_{j}}}\\
					&\leq \sum_{j=0}^{M_{T}-1} \eta^{2} + \frac{L}{(\lambda + L^{2})^{3/2}} \leq \eta^{2}M_{T} + \frac{LM_{T}}{(\lambda + L^{2})^{3/2}}
				\end{align}
				Putting everything together we get:
				\begin{align*}
					\sum_{j=0}^{M_{T}-1} \sum_{t=t_{j}+1}^{t_{j+1}} \| s_{t,a_{t}}\|_{\bar{V}_{j}^{-1}} \leq \sqrt{\frac{5}{4} + C + L^{2}\left( \frac{1}{\lambda} + \frac{1}{\sqrt{\lambda}}\right)}\sqrt{2Td\ln\left(1 + \frac{TL^{2}}{\lambda d} \right)}&\\
					+ \sqrt{2d\ln\left(1 + \frac{TL^{2}}{\lambda d} \right)\left(\eta^{2}M_{T} + \frac{LM_{T}}{(\lambda + L^{2})^{3/2}} \right)}&
                \end{align*}
				Hence the result using the upper bound on $\wt{\beta}(j)$ proved in Lem.~\ref{lem:bound_error_argmax}.
            \end{proof}

            Finally, the last term to bound is \textcircled{c}, that we do similarly to the end of the proof of Lem.\ref{lem:bound_sum_norm_context}.

            \begin{lemma}\label{lem:bound_precision_inverse}
                For all $T\geq 1$,
                \begin{align}
                    \sum_{j=0}^{M_{T}-1} \sum_{t=t_{j}+1}^{t_{j+1}} 2\wt{\beta}(j)\sqrt{\frac{2L}{t_{j}^{3/2}\sqrt{\lambda + L^{2}t_{j}}}} \leq 2\sqrt{2L} M_{T}\beta^{\star}\left[1 + \frac{\eta}{\sqrt{L}}\right]
                \end{align}
                with $\beta^{\star} = 1 + \sqrt{\lambda}S + \sigma\sqrt{d\left(\ln\left(1 + \frac{L^{2}T}{\lambda d}\right) +  \ln\left(\frac{\pi^{2}T^{2}}{6\delta}\right)\right)}$ and $M_{T}$ the number of episodes.
            \end{lemma}
            \begin{proof}{of Lem.~\ref{lem:bound_precision_inverse}.} We have:
                {\small\begin{align}
                    \sum_{j=0}^{M_{T}-1} \sum_{t=t_{j}+1}^{t_{j+1}} 2\wt{\beta}(j)\sqrt{\frac{2L}{t_{j}^{3/2}\sqrt{\lambda + L^{2}t_{j}}}} \leq 2\sqrt{2L}\max_{j} \wt{\beta}(j) \sum_{j=0}^{M_{T}-1} \sqrt{\frac{1}{t_{j}^{3/2}\sqrt{\lambda + L^{2}t_{j}}}}(t_{j+1} - t_{j})
                \end{align}}
                But the condition on the length of the batch ensures that for any batch $j$, $t_{j+1} - t_{j} \leq \eta t_{j} + 1$, thus equation above can be bounded by:
                \begin{align}
                    \sum_{j=0}^{M_{T}-1} \sum_{t=t_{j}+1}^{t_{j+1}} 2\wt{\beta}(j)\sqrt{\frac{2L}{t_{j}^{3/2}\sqrt{\lambda + L^{2}t_{j}}}} &\leq 2\sqrt{2L}\max_{j} \wt{\beta}(j) \left[M_{T} + \sum_{j=0}^{M_{T}-1} \eta\sqrt{\frac{\sqrt{t_{j}}}{\sqrt{\lambda + L^{2}t_{j}}}}\right]\\
                    &\leq 2\sqrt{2L}\max_{j}\wt{\beta}(j) \left[M_{T} + \sum_{j=0}^{M_{T}-1} \frac{\eta}{\sqrt{L}}\right]\\
                    &\leq 2\sqrt{2L}M_{T}\max_{j} \wt{\beta}(j) \left[1+ \frac{\eta}{\sqrt{L}}\right]
                \end{align}
                Hence the result.
            \end{proof}

            Finally, we can finish the proof of Thm.~\ref{thm:regret_efficient_linucb}, but we first recall its statement.

            \begin{theorem*}
                Under Asm.~\ref{assumption:boundness}, for any $\delta>0$ and $T\geq d$, there exists universal constants $C_1,  C_2>0$ such that the regret of \secofulls (Alg.~\ref{alg:low_switching_algo}) is bounded with probability at least $1 - \delta$ by:
                {\small \begin{equation*}
                    \begin{aligned}
                        R_{T} \leq &\, C_{1} \beta^{\star}\left(\sqrt{\left(\frac{5}{4}+ C\right)dT\ln\left(\frac{TL}{\lambda d} \right)} + \frac{L^{3/2}}{\sqrt{\lambda}}\ln(T)\right) + C_2\beta^{\star}M_{T}\max\left\{\sqrt{L} + \eta, \eta^{2} + \frac{L}{\sqrt{\lambda + L^{2}}^{3}}\right\}\\
                    \end{aligned}
                \end{equation*}}
            \end{theorem*}

            \begin{proof}{of Thm.~\ref{thm:regret_efficient_linucb}.} For any $\delta >0$, let's define the event $E$ as in the proof of Lem.~\ref{lem:first_bound_regret}.
                Then conditioned on this event, we have using Lem.~\ref{lem:first_bound_regret}:
                {\small\begin{equation}
                    \begin{aligned}
                        R_{T}(\text{\secofulls}) \leq
                        &\sum_{j=0}^{M_{T}-1} \sum_{t=t_{j}+1}^{t_{j+1}}\frac{4}{t} \left(1 + 2\wt{\beta}(j)\left[\frac{2}{t}
                        + \sqrt{\frac{L}{t_{j}^{3/2}\sqrt{t_{j}L^{2}+ \lambda}}} + L\sqrt{\frac{1}{\lambda} +
                        \frac{1}{\sqrt{\lambda}}} \right] \right)& \\
                        +&\sum_{j=0}^{M_{T}-1} \sum_{t= t_{j}+1}^{t_{j+1}} 2\wt{\beta}(j)\left[  \mathrm{sqrt}_{\mathrm{HE}}\left( s_{t,a_{t}}^\top \text{Dec}_{\textbf{sk}}(\bar{A}_{j}) s_{t,a_{t}} + \frac{L}{t_{j}^{3/2}\sqrt{\lambda + L^{2}t_{j}}}\right) + \frac{1}{t}\right]
                    \end{aligned}
                \end{equation}}
                But using Lem.~\ref{lem:bound_error_argmax} to bound the first term of the RHS equation above but
                also Lem.~\ref{lem:second_bound_regret}, \ref{lem:bound_precision_sqrt}, \ref{lem:bound_sum_norm_context} and
                \ref{lem:bound_precision_inverse} to the bound the second term, we get:
                \begin{equation}
                    \begin{aligned}
                        R_{T}\left( \text{\secofulls}\right) \leq \left( 1 + \ln(T)\right) \Bigg[1 +
                        \beta^{\star}
                        \left(6 + L\sqrt{\frac{1}{\lambda} + \frac{1}{\sqrt{\lambda}}} + \sqrt{\frac{L}{\sqrt{\lambda}}}\right)\Bigg]& \\
                        + \beta^{\star}\sqrt{2d\ln\left(1 + \frac{TL^{2}}{\lambda d} \right)}\Bigg[\sqrt{ \frac{5}{4}+ C + L^{2}\left( \frac{1}{\lambda} + \frac{1}{\sqrt{\lambda}}\right)}\sqrt{T}
                        + \left(\eta^{2} + \frac{L}{(\lambda + L^{2})^{3/2}}\right)M_{T}\Bigg]&\\
                        + 2\sqrt{2L} M_{T}\beta^{\star}\left[1 + \frac{\eta}{\sqrt{L}}\right]&
                    \end{aligned}
                \end{equation}
                with $\beta^{\star} = 1 + \sqrt{\lambda}S + \sigma\sqrt{d\left(\ln\left(1 + \frac{L^{2}T}{\lambda d}\right) +  \ln\left(\frac{\pi^{2}T^{2}}{6\delta}\right)\right)}$ and $M_{T} = 1 + \frac{d\ln\left(1 + \frac{L^{2}T}{\lambda d}\right)}{2\ln\left( \frac{3}{4} + C - \frac{L\eta}{\sqrt{\lambda + L^{2}}} \right)} + \frac{\ln(T)}{\ln(1 + \eta)}$
            \end{proof}

\section{IMPLEMENTATION DETAILS:}

    In this section, we further detail how \secofulls is implemented. In particular,
    we present the matrix multiplication and matrix-vector operations.

    For the experiments, we used the PALISADE library (development version v$1.10.4$) \cite{PALISADE}.
    This library automatically chooses most of the parameters used for the CKKS scheme. In particular the ring dimension
    of the ciphertext space is chosen automatically. In the end, the user only need to choose four parameters: the
    maximum multiplicative depth (here chosen at $100$), the number of bits used for the scaling factor (here $50$),
    the batch size that is to say the number of
    plaintext slots used in the ciphertext (here $8$) and the security
    level (here chosen at $128$ bits for Fig.~\ref{fig:regret_secofulls}).

    \subsection{Matrix/Vector Encoding}\label{app:matrix_encoding}

        Usually, when dealing with matrices and vectors in homomorphic encryption there are multiple ways
        to encrypt those. For example, with a vector $y\in \mathbb{R}^{d}$ one can
        create $d$ ciphertexts encrypting each value $y_{i}$ for all $i\leq d$. This approach is nonetheless expensive
        in terms of memory. An other approach is to encrypt directly the whole vector in a single
        ciphertext. A ciphertext is a polynomial ($X\mapsto \sum_{i=0}^{N} a_{i}X^{i}$) where each coefficient is used to encrypt a value
        of $y$ ($a_{i} = y_{i}$ for $i\leq d$).
        This second method is oftentimes preferred as
        it reduce memory usage.

        It is possible to take advantage of this encoding method in order to facilitate computations,
        \eg matrix multiplication, matrix-vector operation or scalar product. In this work, we
        need to compute the product of square matrices of size $d\times d$, thus we choose to encrypt each matrix/vector as a unique
        ciphertext (assuming $d\leq N$). We have two different encoding for matrices and vectors. For a matrix $A = (a_{i,j})_{i\in \{0, \dots, p-1\}, j\in\{0, \dots, q-1\}}$
        with $p,q \in \mathbb{N}$, we first transform $A$ into a vector of size $pq$,
        $\tilde{a} = (a_{0, 0}, a_{0,1}, \dots, a_{0, q-1}, \dots, a_{1, 0}, \dots, a_{1, q-1}, \dots, a_{p-1,q-1})$.
        This vector is then encrypted into a single ciphertext. But for a vector $y\in \mathbb{R}^{q}$, we create a bigger vector of
        dimension $pq$ (here $p$ is a parameter of the encoding method for vectors),
        $\tilde{y} = (y_{j})_{i\in\{0, \dots, p - 1\}, j\in\{0, \dots, q - 1\}}
         = (y_{0}, \dots, y_{q-1}, y_{0}, \dots, y_{q-1}, \dots, y_{q-1})$. We choose those two encodings
        because the homomorphic multiplication operation of PALISADE only perform a
        coordinate-wise multiplication between two ciphertexts. Therefore, using this encoding, a matrix-vector product
        for a matrix $A\in\mathbb{R}^{p\times q}$, a vector $y\in \mathbb{R}^{q}$, a public key
        $\textbf{pk}$ can be computed as:
        {\small\begin{align*}
            c_{A} \times c_{y} = \text{Enc}_{\textbf{pk}}(\tilde{a}\cdot\tilde{y}) = \text{Enc}_{\textbf{pk}}\Bigg(\Bigg( a_{0,0}y_{0}, a_{0,1}y_{1}, \dots, a_{0, q-1}y_{q-1}, a_{1, 0}y_{0}, a_{1, 1}y_{1}, \dots, a_{1, q-1}y_{q-1},\\
            \dots, a_{p-1, q-1}y_{q-1}\Bigg)\Bigg)&
        \end{align*}}
        with $\tilde{a}$ the encoding of $A$,
        $c_{A} = \text{Enc}_{\textbf{pk}}(\tilde{a})$, $\tilde{y}$ the encoding of $y$ of dimension $pq$ and
        $c_{y} = \text{Enc}_{\textbf{pk}}(\tilde{y})$, $\times$ the homomorphic multiplication operation and $\tilde{a}\cdot\tilde{y}$ the element-wise
        product. Then using $\text{EvalSumCol}$ (an implementation of the $\text{SumColVec}$ method from \cite{han2018efficient} in the
        PALISADE library) to compute partial sums of the coefficients of $c_{A}\times c_{y}$, we get:
        {\small\begin{align*}
            \text{EvalSumCol}(c_{A}\times c_{y}, p, q) = \text{Enc}_{\textbf{pk}}\Bigg(\Bigg( \sum_{j=0}^{q-1} a_{0,j}y_{j}, \dots, \sum_{j=0}^{q-1} a_{0,j}y_{j}, \sum_{j=0}^{q-1} a_{1,j}y_{j}, \dots,
            \sum_{j=0}^{q-1} a_{1,j}y_{j},&\\
             \dots, \sum_{j=0}^{q-1} a_{p-1,j}y_{j}\Bigg)\Bigg)&
        \end{align*}}

        Finally, the matrix-vector product $Ay$ is computed by $\text{EvalSumCol}(c_{A}\times c_{y}, p, q)$
        taking the coefficient at $(j +j\cdot p)_{j\in [p]}$.

    \subsection{Matrix Multiplication}

        Using the encoding of App.~\ref{app:matrix_encoding} we have a way to compute a matrix-vector product
        therefore computing the product between two square matrices $M,N \in \mathbb{R}^{p\times p}$ can be done using a series of
        matrix-vector products. However, this approach requires $p$ ciphertexts to represent a matrix.
        We then prefer to use the method introduced in Sec. $3$ of \cite{jiang2018secure}. This method relies on the following
        identity for any matrices $M,N \in \mathbb{R}^{p\times p}$ and $i,j\in \{0, \dots, p-1\}$:
        \begin{equation}\label{eq:identity_multiplication}
            \begin{aligned}
                (MN)_{i,j} &= \sum_{k = 0}^{p-1} M_{i,k}N_{k,j}\\
                &= \sum_{k = 0}^{p-1} M_{i,[i+k+j]_{p}}N_{[i+k+j]_{p},j}\\
                &= \sum_{k = 0}^{p-1} \sigma(M)_{i,[j+k]_{p}}\tau(N)_{[i+k]_{p},j}\\
                &= \sum_{k = 0}^{p-1} (\phi^{k} \circ \sigma(M))_{i,j}(\psi^{k} \circ \tau(N))_{i,j}
            \end{aligned}
        \end{equation}
        where we define $\sigma, \tau, \psi$ and $\phi$ as:
        \begin{figure}[!h]
        \begin{minipage}{0.45\linewidth}
            \begin{itemize}
                \item $\sigma(M)_{i,j} = M_{i, [i+j]_{p}}$
                \item $\tau(M)_{i,j} = M_{[i+j]_{p}, j}$
            \end{itemize}
        \end{minipage}\hfill
        \begin{minipage}{0.45\linewidth}
            \begin{itemize}
                \item $\psi(M)_{i,j} = M_{i, [i+1]_{p}}$
                \item $\phi(M)_{i,j} = M_{[i+1]_{p}, j}$
            \end{itemize}
        \end{minipage}
        \end{figure}
        and $[.]_{p}$ is the modulo operator. Therefore, using Eq.~\eqref{eq:identity_multiplication} we have that
        computing the product between $M$ and $N$ can simply be done by computing a component-wise multiplication between
        $(\phi^{k} \circ \sigma(M))_{i,j}$ and $(\psi^{k} \circ \tau(N))_{i,j}$ for all $k\in \{0, \dots, p-1\}$.
        Those quantities can in turn be easily computed thanks to a multiplication between a plaintext and a ciphertext (this does not impact the depth of the ciphertext).

    \subsection{Influence of the Security Level}\label{app:security}

        Finally, we investigate the influence of the security level $\kappa$ on the running time and regret of \secofulls. As mentioned
        in Sec.~\ref{sec:discussion} the security parameter $\kappa$ ensures that an attacker has to perform at least $2^{\kappa}$ operations
        in order to decrypt a ciphertext encrypted using an homomorphic encryption scheme. But, the security parameter also has an
        impact on the computational efficiency of our algorithm. Indeed the dimension $N$ of the ciphertext space, \ie the degree of the polynomials in $\mathbb{Z}[X]/(X^{N}+1)$, increases
        with the multiplicative depth $D$ and $\kappa$. However, this means that our algorithm has to compute operations with polynomials of
        higher dimensions hence more computationally demanding.

        \begin{figure}
            \begin{center}
                \includegraphics[width=0.4\linewidth]{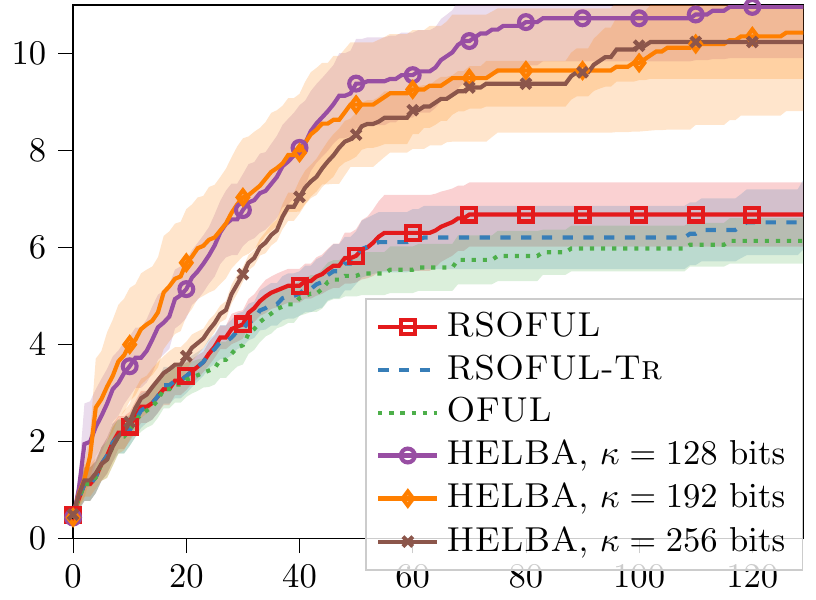}
            \end{center}
            \caption{\label{fig:regret_different_security}Regret of \secofulls for $\kappa\in\{128, 192, 256\}$}
        \end{figure}

        The library PALISADE allows us to choose $\kappa\in\{128, 192, 256\}$. We executed \secofulls with the same parameter and the same environment
        of Sec.~\ref{sec:discussion} except for the parameter $\kappa$ which now varies in $\{128, 192, 256\}$. First, we investigate the regret of for each parameter $\kappa$, this parameter should
        have no impact on the regret \secofulls, as showed in Fig.~\ref{fig:regret_different_security}.

        Second, we investigate the running time for each $\kappa\in\{128, 192, 256\}$. Table~\ref{tab:running_time} shows the ratio between the total computation time
        of $130$ steps using the environment described in Sec.~\ref{sec:discussion} with \secofulls for different security parameters and $\kappa=128$ bits. In order to investigate only the effect of the parameter
        $\kappa$, the results in Table~\ref{tab:running_time} are expressed as a ratio.
        For reference, the total time for $T=130$ steps
        and $\kappa = 128$ bits was {\color{black}$20$ hours and $39$ minutes}. As we observe in Table~\ref{tab:running_time} the impact on the security parameter is around $1$\% and $2$\% of the total computation time for $128$ bits. This
        increase in computation time represents between $20$ and $40$ minutes of computation which in some applications can be prohibitive.

        \begin{table}[t]
            \caption{Ratio of running time for \secofulls as a function of $\kappa$ for the bandit problem of Sec.~\ref{sec:discussion}. We use the running time
            with $\kappa = 128$ bits and $T = 130$ steps as a reference to compute the ratio between this time and the total time for $\kappa\in \{192, 256\}$. }
            \label{tab:running_time}
            \vskip 0.15in
            \begin{center}
            \begin{small}
            \begin{sc}
            \begin{tabular}{lcccr}
            \toprule
            $\kappa$ (Bits) & Ratio Execution Time  \\
            \midrule
            $128$ & $1$\\
            $192$    & $1.016$\\
            $256$    & $1.026$\\
            \bottomrule
            \end{tabular}
            \end{sc}
            \end{small}
            \end{center}
            \vskip -0.1in
            \end{table}


\end{appendix}
\end{document}